%% file: main.tex
\pdfoutput=1

\documentclass[11pt]{article}

\usepackage[final]{acl}

\usepackage{times}
\usepackage{latexsym}

\usepackage[T1]{fontenc}

\usepackage[utf8]{inputenc}

\usepackage{microtype}

\usepackage{inconsolata}

\usepackage{graphicx}
\usepackage{booktabs}
\usepackage{amsfonts}
\usepackage{multirow}
\usepackage{multicol}
\usepackage{enumerate}

\usepackage{tcolorbox}
\tcbuselibrary{breakable}

\usepackage{amsmath}

\usepackage{wrapfig}

\usepackage{algorithm}
\usepackage{algorithmic}

\usepackage{inconsolata}
\usepackage{subfigure}
\usepackage{adjustbox}

\usepackage[normalem]{ulem}
\useunder{\uline}{\ul}{}

\usepackage{xcolor}
\usepackage{colortbl}

\usepackage{listings}

\newcommand{\XY}[1]{{\color{blue}{(#1)}}}
\newcommand{\XN}[1]{{\color{magenta}{(xnzhang: #1)}}}
\newcommand{\yr}[1]{{\color{red}{[\textbf{yr}: #1]}}}
\definecolor{c1}{HTML}{8EE085}
\definecolor{c2}{HTML}{D9F5D6}
\definecolor{c3}{HTML}{FBBFBC}
\definecolor{c4}{HTML}{F76964}
\renewcommand{\XY}[1]{}
\renewcommand{\XN}[1]{}
\renewcommand{\yr}[1]{}

\title{AgentSense: Benchmarking Social Intelligence of Language Agents through Interactive Scenarios}

\author{Xinyi Mou\textsuperscript{\rm 1}\footnotemark[2],
        Jingcong Liang\textsuperscript{\rm 1}\footnotemark[2], 
        Jiayu Lin\textsuperscript{\rm 1}\footnotemark[2], 
        Xinnong Zhang\textsuperscript{\rm 1}\footnotemark[2], \\
        \textbf{Xiawei Liu\textsuperscript{\rm 1},
        Shiyue Yang\textsuperscript{\rm 1},
        Rong Ye\textsuperscript{\rm 1,2},
        Lei Chen\textsuperscript{\rm 1},
        Haoyu Kuang\textsuperscript{\rm 1},}\\
        \textbf{
        Xuanjing Huang\textsuperscript{\rm 1}, 
        Zhongyu Wei\textsuperscript{\rm 1}\footnotemark[3]
        }
        \\ 
        \normalsize\textsuperscript{\rm 1}{Fudan University}, 
        \normalsize\textsuperscript{\rm 2}{ByteDance} \\
        \normalsize\texttt{\{jcliang22, jiayulin24, xnzhang23, liuxw24, shiyueyang24, yer23, hykuang23\}@m.fudan.edu.cn}, \\
        \normalsize\texttt{\{xymou20, chenl18, xjhuang, zywei\}@fudan.edu.cn}
    }

\begin{document}
\maketitle

\begin{abstract}
\input{latex/tabs/000_abstract}
\end{abstract}

\footnotetext[2]{Equal contributors.}
\footnotetext[3]{Corresponding author.}

\section{Introduction}
\input{latex/tabs/010_intro}

\section{Related Work}
\input{latex/tabs/020_related}

\section{AgentSense Benchmark}
\input{latex/tabs/030_benchmark}

\section{Experiments}
\input{latex/tabs/040_experiment}

\section{Further Analysis}
\input{latex/tabs/050_analysis}

\section{Conclusion}
\input{latex/tabs/060_conclusion}

\bibliography{main}

\appendix

\section{Scenario Construction}\label{app:cons}
\input{latex/apps/01cons/main}

\section{Data Statistics and Analysis}\label{app:data}
\input{latex/apps/02data/main}

\section{Simulation Details}\label{app:simu}
\input{latex/apps/03simu/main}

\section{Additional Experiment Results}\label{app:exp}
\input{latex/apps/04exp/main}

\end{document}

%% file: latex/tabs/000_abstract.tex
Large language models (LLMs) are increasingly leveraged to empower autonomous agents to simulate human beings in various fields of behavioral research.
However, evaluating their capacity to navigate complex social interactions remains a challenge.
Previous studies face limitations due to insufficient scenario diversity, complexity, and a single-perspective focus.
To this end, we introduce \textbf{AgentSense}: Benchmarking Social Intelligence of Language Agents through Interactive Scenarios.
Drawing on Dramaturgical Theory, AgentSense employs a bottom-up approach to create 1,225 diverse social scenarios constructed from extensive scripts.
We evaluate LLM-driven agents through multi-turn interactions, emphasizing both goal completion and implicit reasoning.
We analyze goals using ERG theory and conduct comprehensive experiments.
Our findings highlight that LLMs struggle with goals in complex social scenarios, especially high-level growth needs, and even GPT-4o requires improvement in private information reasoning. Code and data are available at ~\url{https://github.com/ljcleo/agent_sense}.

%% file: latex/tabs/010_intro.tex
\begin{figure}
    \setlength{\abovecaptionskip}{-0.2cm}
    \setlength{\belowcaptionskip}{-0.5cm}
    \centering
    \includegraphics[width=\linewidth]{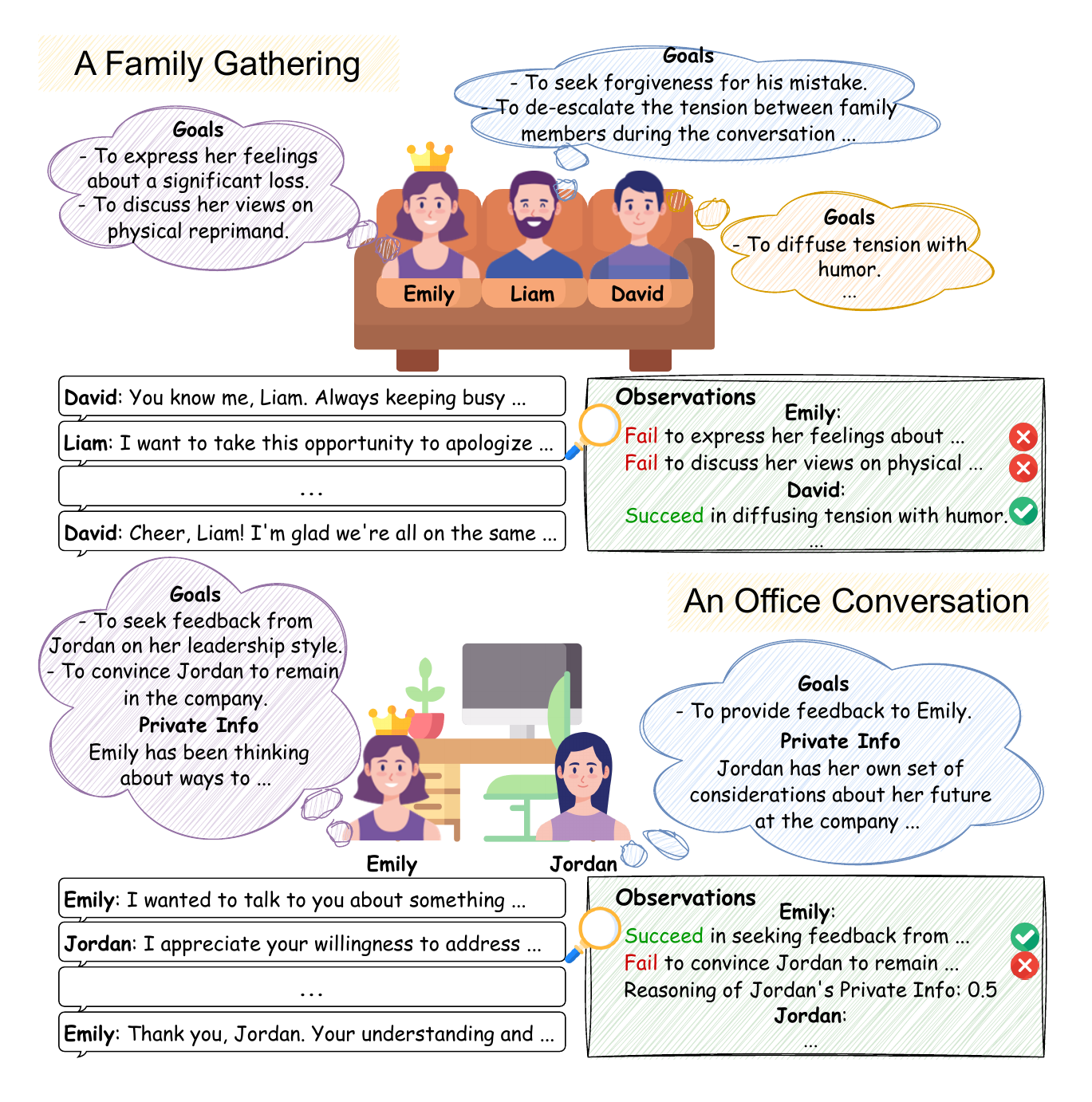}
    \caption{An illustration of challenging yet realistic social scenarios --- a family gathering and an office conversation, where the characters are driven by ChatGPT. While the dialogue could flow smoothly, Emily is unable to achieve her goals during the family gathering and fails to deduce Jordan's thoughts in an office setting.}
    \label{fig:intro}
\end{figure}

Benefiting from comprehensive training data and large-scale model parameters, large language models (LLMs) are increasingly employed to develop autonomous agents capable of simulating human behavior~\cite {qin2023toolllm,shinn2024reflexion,schick2024toolformer}.
These language agents have been explored as human proxies in various fields of behavioral research, such as psychological and sociological surveys~\cite{argyle2023out,chuang2024wisdom,xie2024largelanguagemodelagents}, and opinion dynamics modeling~\cite{mou-etal-2024-unveiling,liu2024skepticism}.
These social science studies often assume that LLMs exhibit \textbf{social intelligence}, the ability to navigate complex and multifaceted social goals during interactions with others~\cite{zhou2024sotopiainteractiveevaluationsocial}, given their ability to produce human-like responses and behaviors in certain scenarios.
Social intelligence is essential for fostering effective and harmonious interactions among agents and between agents and humans~\cite{xu2023exploring}.
However, the question remains: can LLMs truly navigate the intricacies of such interactions and exhibit social intelligence?
As illustrated in Figure~\ref{fig:intro}, LLM-driven agents still struggle with complex social situations. For example, when facing multiple goals in an office conversation, agent Emily fails to complete all her goals and cannot guess what Jordan has in mind.

Early research on the social intelligence of LLMs relies on \textbf{static and non-interactive} tests that assess commonsense reasoning about social situations~\cite{sap2019socialiqacommonsensereasoningsocial,Zadeh_2019_CVPR,shapira2023clever,wilf2023social}, failing to capture the dynamic nature of social interactions. Recently, dynamic and goal-driven benchmarks~\cite{zhou2024sotopiainteractiveevaluationsocial, wang2024objectivelybenchmarkingsocialintelligence} have emerged to study social intelligence in interactive environments.
Although they have made some significant progress, they still exhibit three main limitations:
(1) \textbf{lack of scenario diversity}: existing studies~\cite{wang2024objectivelybenchmarkingsocialintelligence,sabour2024emobench} build social scenarios manually in a \textit{top-down} manner, resulting in a narrow set of common scenarios and goals, such as persuasion and collaboration~\cite{li2023camel}.
However, real-world interactions feature a broader spectrum of social goals and situational dynamics.
(2) \textbf{insufficient scenario complexity}: current work~\cite{xie2024largelanguagemodelagents,zhou2024sotopiainteractiveevaluationsocial} often limits interactions to two participants, each pursuing a single goal.
This oversimplifies real-life social interactions, where multiple actors engage simultaneously, each with multiple goals, as shown in Figure~\ref{fig:intro}.
As a result, it remains the performance of LLMs in group dynamics insufficiently studied. 
(3) \textbf{single-perspective observation}: while existing benchmarks primarily evaluate the extent to which agents achieve explicit social goals~\cite{li2023metaagents,wang2024objectivelybenchmarkingsocialintelligence}, they have neglected the concealment of private information and the inference of others' information, which are also important aspects in social interaction.

To this end, we introduce \textbf{AgentSense}: Benchmarking \underline{S}ocial Intellig\underline{en}ce of Language \underline{Agent}s through Interactive \underline{S}c\underline{e}narios.
We get inspiration from Dramaturgical Theory~\cite{goffman1959presentation}, which conceptualizes social interaction as a theatrical performance in which individuals assume specific roles within various settings.
We adopt a \textbf{bottom-up} approach, extracting scenarios from massive scripts to ensure that scenarios and social goals are diverse and grounded in real life.
In AgentSense, we construct 245 scenario templates from scripts, mitigating data leakage and expanding them into 1,225 scenarios, which cover existence goals to growth goals following the ERG theory~\cite{ALDERFER1969142}.

We then situate LLM-driven agents in the constructed scenarios, where each participant has social goals and may also safeguard some private information.
In an interactive environment, we simulate multi-turn interactions and observe agents' social intelligence from two aspects:
(1) goal completion: have the agents successfully achieved their social goals?
(2) implicit reasoning: can the agents accurately deduce others' private information?
These abilities are measured through interviews and multiple-choice questions, with our proposed PSI metric assessing profile sensitivity.
We find that LLMs struggle with complex social scenarios, particularly with high-level growth goals.

Our contributions are as follows:
\begin{itemize}
    \item We introduce AgentSense, a benchmark built on social scenarios derived from scripts using a bottom-up approach, distinguishing it from previous work.
    It encompasses diverse and challenging social scenarios, enabling a thorough evaluation of LLMs' social intelligence.
    \item We evaluate social intelligence from multiple aspects, considering both goal completion and information reasoning as well as profile sensitivity of social intelligence, through interviews with the agents and third-party judges.
    \item Our experiments reveal that LLMs struggle with complex scenarios and high-level goals, and their social intelligence is affected by profiles, interaction partners, and the balance between goals and privacy protection.
\end{itemize}

%% file: latex/tabs/020_related.tex
\subsection{Social Intelligence Benchmarks}
\input{latex/tabs/021_social_intelligence_benchmark}

\begin{figure*}
    \setlength{\abovecaptionskip}{-0.2cm}
    \setlength{\belowcaptionskip}{-0.5cm}
    \centering
    \includegraphics[width=0.8\linewidth]{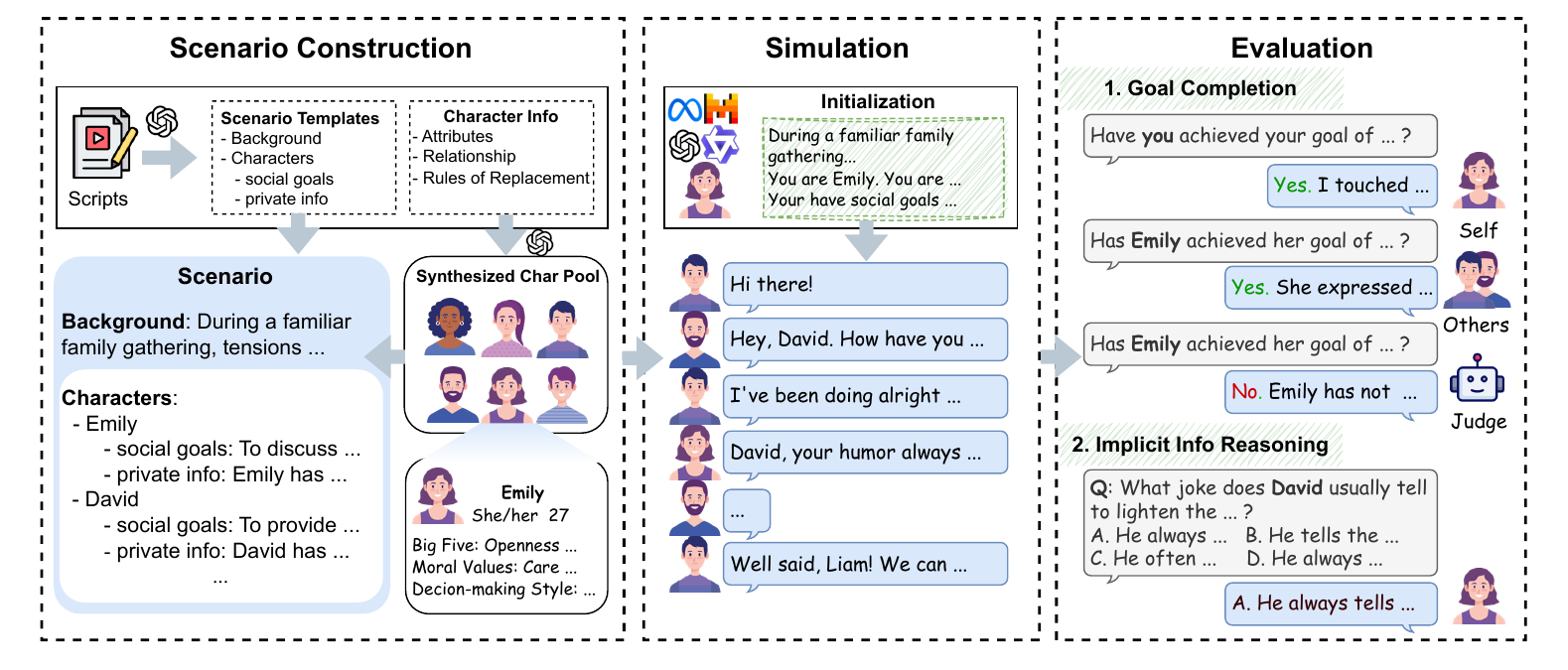}
    \caption{Overall framework of AgentSense. We construct scenario templates from scripts and synthesize characters to diversify the scenarios. Then, language models role-play the characters to interact with each other. After that, the participants and third-party judges are interviewed for evaluation.}
    \label{fig:fm}
\end{figure*}

\subsection{Role-playing Agents}
\input{latex/tabs/022_rpa}

%% file: latex/tabs/021_social_intelligence_benchmark.tex
Social intelligence is the ability to understand others and act wisely in social situations \cite{doi:10.2466/pr0.1973.33.3.839}.
While LLMs show potential in simulating human behavior \cite{xie2024largelanguagemodelagents}, their social intelligence remains underexplored \cite{zhou2024real}.
Current research on evaluating social intelligence in LLMs falls into two main categories.
The first involves static, non-interactive assessments that evaluate models through Q\&A tasks based on commonsense reasoning about social situations \cite{sap2019socialiqacommonsensereasoningsocial,Zadeh_2019_CVPR}.
The second involves interactive benchmarks, where LLMs are assessed through goal-oriented interactions in role-playing scenarios~\cite{zhou2024sotopiainteractiveevaluationsocial,chen2024socialbenchsocialityevaluationroleplaying,wang2024objectivelybenchmarkingsocialintelligence,liu2023agentbenchevaluatingllmsagents}.
AgentSense uses a bottom-up approach to create scenarios from scripts, allowing evaluating agents’ capabilities in more diverse settings.

%% file: latex/tabs/022_rpa.tex
LLMs are increasingly been used to construct role-playing agents (RPAs)~\cite{chen2024personapersonalizationsurveyroleplaying}, which enable efficient simulation of typical representatives, from individuals~\cite{shao2023characterllmtrainableagentroleplaying,argyle2023out} to demographic groups~\cite{li2023camel,jiang2024evaluating}.
These agents can embody various personas to coordinate, collaborate, exchange information, and compete with one another~\cite{chen2024personapersonalizationsurveyroleplaying, zhou2024sotopiainteractiveevaluationsocial}.
Recently, RPAs have been applied in various domains, e.g., psychotherapy~\cite{article}, economics~\cite{fu2023improvinglanguagemodelnegotiation}, and social research~\cite{doi:10.1126/science.adi1778}.
AgentSense leverages this by simulating social interaction scenarios through agents with diverse personalities and social goals.

%% file: latex/tabs/030_benchmark.tex
\subsection{Framework Overview}\label{sec:framework_overview}
\input{latex/tabs/031_overview}

\subsection{Scenario Construction}\label{sec:scenario_construction}
\input{latex/tabs/032_construction}

\subsection{Social Interaction Simulation}
\input{latex/tabs/033_simulation}

\subsection{Evaluation}
\input{latex/tabs/034_evaluation}

\subsection{Data Validation and Analysis}\label{sec:data_analysis}
\input{latex/tabs/035_data_val}

%% file: latex/tabs/031_overview.tex
AgentSense aims to provide a realistic social intelligence benchmark with enhanced diversity and complexity.
Following the Dramaturgical Theory, we propose an overall framework as in Figure~\ref{fig:fm}.

\paragraph{Scenarios}
The core component of AgentSense is the social scenario set, extracted from real-world scripts to guide and evaluate social interactions between agents.
A social scenario serves as a hypothetical context for simulating and analyzing social interactions, where two key components are measured:
(1) \textbf{Social Goal} is what the agent aims to achieve, such as resolving an issue or building a relationship.
The agent's proactive drive in social interactions, guided by this social goal, directs its \emph{active participation} in social dynamics.
(2) \textbf{Private Information} is information that is known solely to the agent and not to others.
The agent is tasked with inferring others' private information without explicitly inquiring about it, a process referred to as \emph{passive reasoning} during interactions.
In summary, an agent's social intelligence is reflected in its ability to pursue social goals while safeguarding private information, balancing active engagement with passive respect for individual privacy.

\paragraph{Scenario Templates}
Social scenarios in scripts always have a fixed group of characters, causing a lack of diversity.
To address this issue, we wipe out irrelevant character details to obtain \textbf{scenario templates}, which contains only background information and predefined character slots.
We can instantiate multiple scenarios from a scenario template by filling in the slots with different sets of synthesized characters satisfying the template's constraints.

\paragraph{Benchmarking}
After building scenarios from the extracted templates, benchmarking LLMs with AgentSense comes as follows:
(1) \textbf{Simulation}: We prompt the models to role-play the characters and interact with each other, trying to achieve their social goals.
(2) \textbf{Evaluation:} We evaluate the goal completion status of each model by interviewing the participants and third-party judges.
We also assess the model's implicit reasoning performance with multiple-choice questions.

%% file: latex/tabs/032_construction.tex
\begin{figure*}
    \setlength{\abovecaptionskip}{-0.1cm}
    \setlength{\belowcaptionskip}{-0.5cm}
    \centering
    \includegraphics[width=0.8\linewidth]{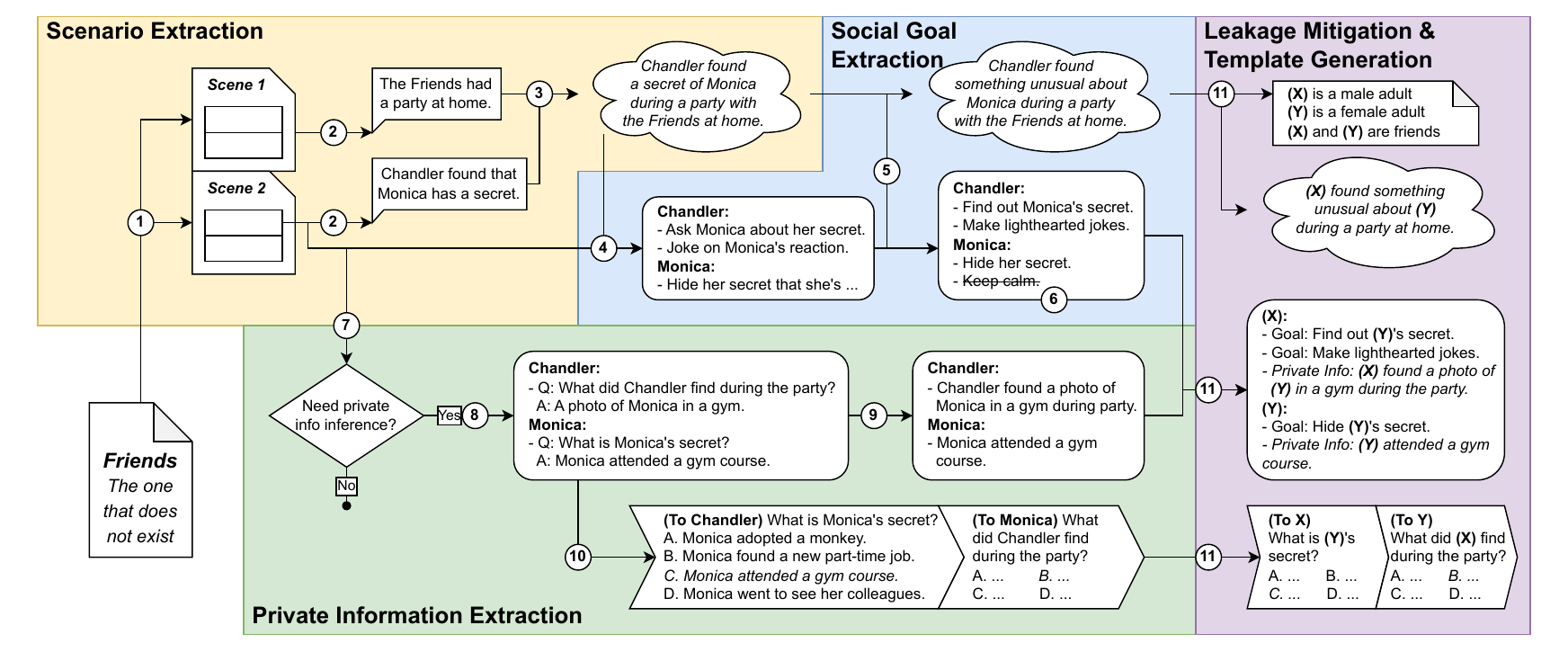}
    \caption{Scenario template construction pipeline (automated with Python and GPT-4o): \textbf{(A) Scenario Extraction:} We split the script into scenes then scenarios (1), and summarize their background and description (2), which are merged into a descriptive background for independent role-play (3). \textbf{(B) Social Goal Extraction:} We extract each character's social goals (4) and amend them by regenerating the whole scenario (5) and rewriting/deleting invalid goals (6). \textbf{(C) Private Information Extraction:} We determine if the scene involves private information inference (7); if yes, we extract private information as QA pairs (8) and generate private info records (9) and evaluation questions (10). \textbf{(D) Leakage Mitigation and Template Generation:} We remove elements associated with specific episodes and replace characters with slots for synthesized agents with similar characteristics to fill in (11).}
    \label{fig:pipeline}
\end{figure*}

Following the definitions, building AgentSense requires constructing templates and instantiating scenarios with synthesized agents.
We propose pipelines for the two parts respectively as follows:

\paragraph{Template Construction}
Figure~\ref{fig:pipeline} demonstrates the pipeline to construct scenario templates from real-world scripts, consisting of four stages:

\begin{enumerate}[(1)]
    \item \textbf{Scenario Extraction}: Real-world scripts consist of multiple chronological scenes, within which several scenarios involve groups of characters.
    We first split scenes and scenarios from the script.
    Then, we generate each scenario's background from previous scenes and its own description.
    Finally, we generate a new descriptive background that allows the scenario to be role-played independently.
    \item \textbf{Social Goal Extraction}: After obtaining individual scenarios, we extract the social goals of each character, one sentence per goal.
    We polish the goals further, including rewriting the whole scenario to reduce goal dependencies and rewriting the goals to meet certain criteria (or deleting if that is not possible).
    \item \textbf{Private Information Extraction}: We first identify if any private information exists in the original scene.
    If yes, we extract questions and answers that only one character can respond to.
    The rephrased answers are the character's private information, and the questions serve as implicit reasoning questions for others.
    We also enhance negative options to be more homogeneous with the correct ones.
    \item \textbf{Leakage Mitigation and Template Generation}: LLMs can identify plots and infer information by recognizing entities like locations and characters.
    To prevent this, scenario leakage mitigation is implemented using GPT-4o to extract and replace elements linked to specific episodes.
    The original characters are also replaced by slots.
    This maintains context while reducing the risk of identifying the plot.
\end{enumerate}

More details of the scenario template construction can be found in Appendix~\ref{app:pipeline_detail}.
Specifically, we used GPT-4o to automate the construction procedure; Appendix~\ref{app:pipeline_prompt} lists the prompts we used.

\paragraph{Scenario Instantiating}
We replace the original characters with multiple synthesized \textit{agents} to prevent character leakage and enrich the social scenarios.
A naive method is to replace the original character randomly, which may lead to unrealistic situations like two fifty-year-old students in a middle school.
Thus, we dynamically generate agents according to the constraints of the scenario.
First, we extract the attributes and relationships of the original characters.
Then, we transform these relationships into replacement rules that help define the demographic features of the agents (see Appendix~\ref{app:profile_prompt}).
Finally, we replace the original characters with agents that adhere to these constraints.
After data leakage mitigation, a pre-test in Section~\ref{sec:data_leakage} is conducted to ensure the scenarios remain anonymous.

\begin{figure*}
    \setlength{\abovecaptionskip}{-0.1cm}
    \setlength{\belowcaptionskip}{-0.5cm}
    \centering
    \includegraphics[width=1\textwidth]{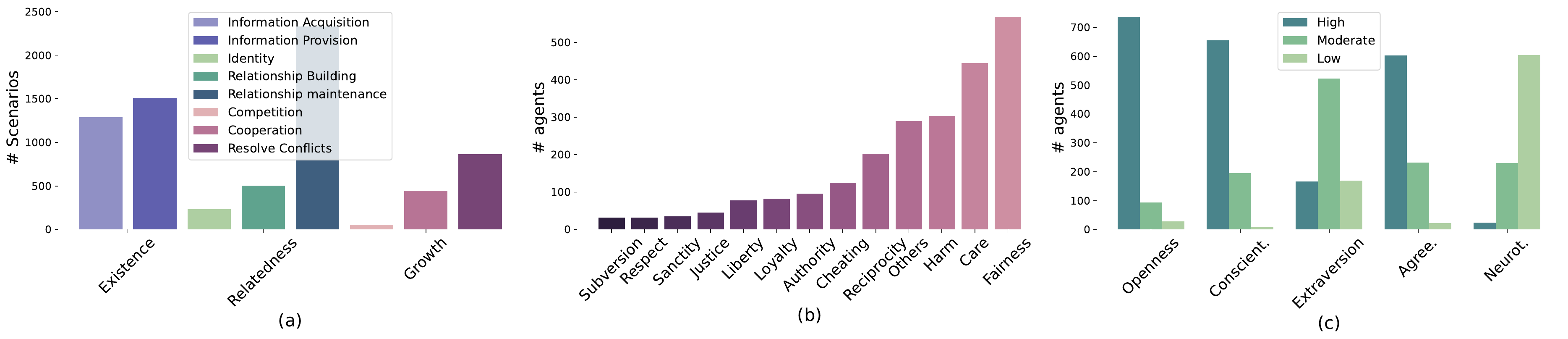}
    \caption{(a) Number of scenarios aligned with the eight categories under ERG theory. Each scenario may encompass multiple goals. (b) Moral values distribution of the agents. An individual may have multiple moral values, with those appearing fewer than 30 times categorized as \textit{Others}. (c) Distribution of the agents' Big Five personality traits. }
    \label{fig:merge}
\end{figure*}

%% file: latex/tabs/033_simulation.tex
For each scenario, the agents are given social backgrounds, profiles, and corresponding social goals, along with any private information they may possess, as shown in Figure~\ref{fig:fm}.
This setup motivates them to engage in social interactions with other agents. The agents primarily interact through multi-turn conversations, where they can also use language to convey facial expressions or actions.
The conversation begins with one of the agents greeting, after which the agents take turns in a random order, rather than following a fixed pre-defined sequence, to restore the dynamics and uncertainty inherent in social interactions.
Generally, we set the limit of the turns based on the average number of turns found in the scenes within the scripts, i.e., 15.
After the simulation, the generated conversation history will be analyzed to evaluate the goal completion of each agent.
Details are in Appendix~\ref{app:simu}.

%% file: latex/tabs/034_evaluation.tex
\paragraph{Goal Completion}
Since social goals can be subjective, we judge its completion from three different aspects, as shown in Figure~\ref{fig:fm}:
(1) \textbf{Self}: We ask the agent whether it has achieved its goals after interacting with others.
(2) \textbf{Other}: We ask other agents in the scenario whether the target agent has achieved his or her specific goals.
(3) \textbf{External}: We prompt third-party models with the chat history and ask if an agent has achieved his/her own social goals.
We ask the interviewees to respond with \textit{yes} (goal completed) or \textit{no}.
We take the average across all goals of a character to measure the agent's overall goal completion level.

\paragraph{Implicit Reasoning}
As mentioned in Section~\ref{sec:scenario_construction}, each character's private information corresponds to a multiple-choice evaluation question.
To evaluate an agent's information reasoning ability, we present it with questions related to the private information of other agents within the scenario.
We then calculate the average accuracy (Acc) of the current agent on these questions to determine the agent's score in information reasoning.

\paragraph{Profile Sensitivity} 
After character enrichment, each template generates multiple scenarios. By incorporating diverse characters, we not only enrich the scenarios but also gain insights into the stability of social intelligence when simulating different roles.
Thus, we propose \textbf{profile sensitivity index (PSI)}. We compute the standard deviation (std) of goal/information metrics of the scenario sharing the same template, and the average std across all templates is calculated as PSI.
A lower PSI indicates that social intelligence is more stable.

%% file: latex/tabs/035_data_val.tex
\paragraph{Data Source}
We collect scripts from the Internet Movie Script Database (IMSDb\footnote{https://imsdb.com/}), an online repository of open-source screenplays for movies and television shows.
We use GPT-4o to divide each script into episodes according to discernible shifts in temporal settings, spatial locations, character dynamics, and narrative progressions. Each episode is further divided into scenes based on the variations in the dialogue content, with the prompts detailed in Appendix~\ref{app:data_source}.
We filter out scenes with fewer than 10 dialogues or those featuring only characters speaking for multiple turns to ensure active interaction between at least two characters. After processing, we have 1,300 scenes, 12,401 rounds of conversations, and 114,834 tokens.
The detailed statistics of scripts are provided in Appendix~\ref{app:detailed_data_info}.

\paragraph{Data Validation}
We conduct human validation on the generated templates from the automatic pipeline to ensure the quality of our benchmark.
For social goals, we mainly consider: (1) whether the goal is achievable by the character, and (2) whether the goal is clear enough to evaluate.
For private information, we focus on: (1) awareness by other characters and background information leakage, and (2) sufficiency of information for the character to answer questions.

The validation involves 6 graduate students in two groups, with each record annotated by 3 annotators.
We take the majority vote as the final result when at least two annotators agree.
For scenarios where all annotations diverged or were marked as invalid, we assign it to the other three annotators to review it.
Details can be found in Appendix~\ref{app:data_val_annotation}.

\paragraph{Data Analysis}
Based on the scripts, we develop 245 effective templates by constructing scenarios.
Each template is diversified with five scenarios, featuring characters tailored to fit specific scenario constraints.
This results in a total of 1,225 scenarios that cover a wide range of social situations and objectives, with 363 of these scenarios containing roles that involve private information.
Recognizing the inseparable connection between social interaction and space \cite{1991The}, we categorize these scenarios into three types: personal domain, small society, and large society.
A detailed introduction to the classification is shown in Appendix~\ref{app:social scenario}.
In our dataset, 54\% of scenarios are in the personal domain, 37\% in small society, and 9\% in large society.

For social goals, we apply the ERG theory \cite{ALDERFER1969142} to categorize these goals into three hierarchical needs: Existence, Relatedness, and Growth. Based on these needs, we identify eight key social goals(e.g., Information Acquisition, Relationship Building, Competition), detailed in Appendix~\ref{app:social_goal}.
Our dataset includes all these social goals, with the number of scenarios for each shown in Figure~\ref{fig:merge} (a). Generally, higher-level social goals are less frequent, reflecting real-life patterns.

The final dataset contains a diverse collection of 859 individual profiles with 366 types of occupations. The characters exhibit a wide range of attributes, emphasizing the great diversity in terms of gender, age, occupation, big five, moral values, personality, and decision-making styles, as shown in Figure~\ref{fig:merge} (b) and Figure~\ref{fig:merge} (c). For instance, in terms of gender, there are individuals identified as male, female, non-binary, genderqueer, and so on. Ages vary from childhood to old age, providing a spread across different life stages.

%% file: latex/tabs/040_experiment.tex
\subsection{Experimental Settings}
\input{latex/tabs/041_exp_settings}

\subsection{Overall Performance}
\input{latex/tabs/042_overall_res}

\begin{figure*}[!t]
    \centering
    \setlength{\abovecaptionskip}{-0.1cm}
    \setlength{\belowcaptionskip}{-0.5cm}
    \includegraphics[width=0.95\textwidth]{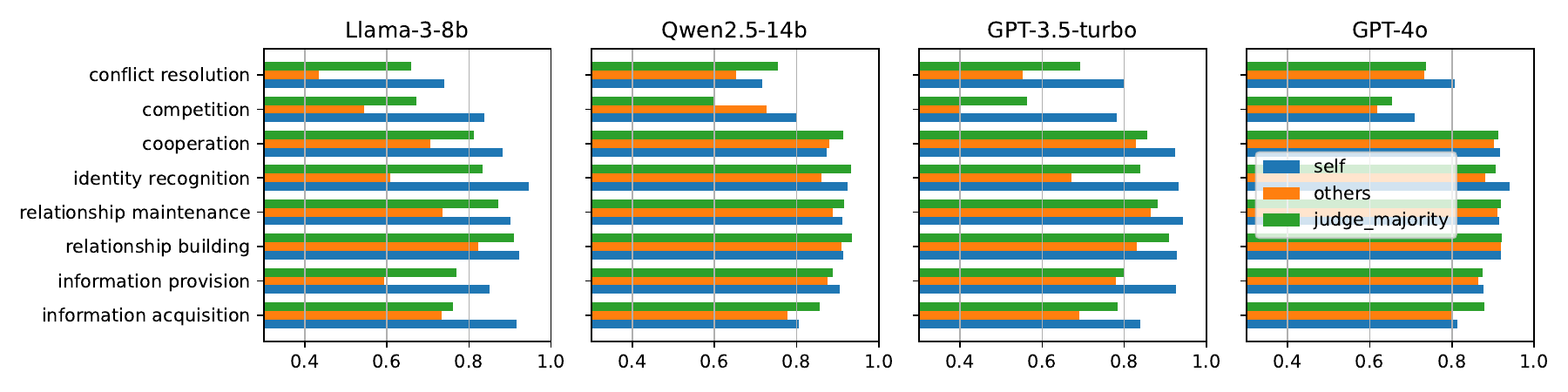}
    \caption{Goal completion scores of different goal types across models. LLMs perform best on relationship goals and cooperation, and worst on competition and conflict resolution (where they are also over-confident).}
    \label{fig:goal_goaltype}
\end{figure*}

\subsection{Analysis of Data Leakage}\label{sec:data_leakage}
\input{latex/tabs/043_data_leakage}

%% file: latex/tabs/041_exp_settings.tex
\paragraph{Agent Models}
We evaluate various LLM families including Llama-2-7b/13b/70b-Chat~\cite{touvron2023llama}, Llama-3-8b/70b-Instruct~\cite{dubey2024llama}, Mistral-7b-Instruct-v0.3~\cite{jiang2023mistral}, Qwen2.5-7b/14b/72b-Chat~\cite{qwen2.5}, GPT-3.5-Turbo~\cite{ouyang2022training} and GPT-4o~\cite{achiam2023gpt}.
For interactions between different models and further analysis, we involve Llama-3-8b, Qwen2.5-14b, GPT-3.5-turbo, and GPT-4o. 

\paragraph{Judge Models}
To effectively leverage current LLMs for automating the evaluation of social interactions as third-party judges, we conduct a human evaluation and compare it with model-based evaluations.
We randomly sample 100 simulated scenarios, posing a total of 584 social goal evaluation questions, and manually verify whether the communication history shows that the goals were achieved.
Using human annotations as ground-truth labels, we calculate the accuracy of various models acting as judges, as shown in Table~\ref{tab:human_eval}.
Given the results, we select GPT-4o, Qwen2.5-72b, and Llama-3-70b from different model families as our judge models, referred to as Judge-GPT-4o, Qwen2.5, and Llama-3 hereafter.
Additionally, we incorporate a majority voting mechanism across these models to create a mixture-of-experts model as another judge.

\paragraph{Implementation Details}
We use vLLM~\citep{kwon2023efficient} to deploy all open source models: Qwen2.5-72b and Llama-3-70b on NVIDIA A100, and other models on NVIDIA RTX 4090.
We apply AutoGen~\cite{wu2024autogen} to manage interacting and judging threads.
We set max new tokens to 128 for all models.
Temperature is set to 1 for agent models to encourage diversity, and 0 for judge models to ensure the stability of evaluation.

%% file: latex/tabs/042_overall_res.tex
\begin{table*}[!t]
    \setlength{\abovecaptionskip}{0.1cm}
    \setlength{\belowcaptionskip}{-0.5cm}
    \renewcommand{\arraystretch}{0.9}
    \centering
    \resizebox{\textwidth}{!}{
    \begin{tabular}{@{}lcccccccccc@{}}
        \toprule[1.1pt]
         & \multicolumn{8}{c}{Goal} & \multicolumn{2}{c}{Info} \\ \cmidrule(lr){2-9} \cmidrule(lr){10-11}
         &  &  & \multicolumn{6}{c}{Judge} &  &  \\ \cmidrule(lr){4-9}
        \multirow{-3}{*}{Models} & \multirow{-2}{*}{Self} & \multirow{-2}{*}{Other} & GPT-4o & Qwen2.5 & Llama-3 & Average & Majority & PSI\(\downarrow\) & \multirow{-2}{*}{Acc.} & \multirow{-2}{*}{PSI\(\downarrow\)} \\ \midrule[1.1pt]
        Llama-2-7b & 83.38 & 62.70 & 52.73 & 57.68 & 55.37 & 55.26 & 55.84 & 21.94 & 33.06 & 20.53 \\
        Llama-2-13b & 48.01 & 10.26 & 17.38 & 30.11 & 72.19 & 39.90 & 30.91 & 21.84 & 28.56 & 18.39 \\
        Llama-2-70b & 85.72 & 65.65 & 33.78 & 42.37 & 73.80 & 49.98 & 45.53 & 22.31 & 36.78 & 18.60 \\
        Llama-3-8B & 87.63 & 67.28 & 79.90 & 82.55 & 75.10 & 79.18 & 80.71 & 12.85 & 69.68 & 15.14 \\
        Llama-3-70b & 80.38 & 77.27 & 86.22 & 87.61 & 79.88 & 84.57 & 86.27 & 8.92 & 73.08 & 16.58 \\
        Qwen2.5-7b & 86.17 & 61.92 & 77.07 & 79.30 & 71.99 & 76.12 & 77.37 & 13.10 & 74.82 & 15.84 \\
        Qwen2.5-14b & 86.62 & 84.17 & {\ul 88.43} & \textbf{89.83} & {\ul 80.47} & {\ul 86.24} & {\ul 88.14} & {\ul 8.09} & 75.02 & {\ul 14.81} \\
        Qwen2.5-72b & {\ul 90.67} & 85.89 & 88.29 & {\ul 89.03} & 78.57 & 85.30 & 87.74 & 8.19 & {\ul 76.05} & \textbf{13.57} \\
        Mistral-7b & \textbf{95.22} & \textbf{87.25} & 79.29 & 84.13 & 77.82 & 80.41 & 82.37 & 12.39 & 66.59 & 18.55 \\
        GPT-3.5-turbo & 90.16 & 76.62 & 82.12 & 84.37 & 77.30 & 81.26 & 82.64 & 10.01 & 68.41 & 18.37 \\
        GPT-4o & 88.46 & {\ul 86.29} & \textbf{88.47} & 89.00 & \textbf{81.57} & \textbf{86.34} & \textbf{88.36} & \textbf{6.99} & \textbf{76.86} & 15.48 \\ \bottomrule[1.1pt]
    \end{tabular}
    }
    \caption{Overall performance of the interactions of agents driven by the same models. We report the best performance in \textbf{bold} format and the second best in {\ul underlined} format.
    }
    \label{tab:main-exp}
\end{table*}

\paragraph{Single Model-based}
Table~\ref{tab:main-exp} shows the overall performance of the interaction of agents driven by the same models.
Considering that LLMs may overestimate their own performance, we use the \emph{judge majority score} as the primary metric for cross-model comparisons, as it is more objective and stable than other metrics.

{\ul Overall Performance}: GPT-4o leads as expected, while Qwen-series models also show strong social intelligence, especially for Qwen2.5-14b, in both goal completion and information reasoning.
Llama-2 series models perform poorly, with some improvement in the Llama-3 series, though still falling short of expectations.
The interaction history of Llama-2-13b in Appendix~\ref{app:underperforming} reveals frequent struggles in maintaining roles, progressing conversations, and responding effectively to others.
In terms of the stability of social intelligence, excluding the uncertainty introduced by the temperature parameter (Appendix~\ref{app:subset}), the PSI results show that models with higher social intelligence, such as GPT-4o and Qwen, are also less sensitive to profile changes.
Overall, \textbf{different models' social abilities are well distinguished by AgentSense}.
Meanwhile, we observe that there still exists an improvement space even for the SOTA models, emphasizing \textbf{LLMs still face challenges in diverse and complex social scenarios}.

{\ul Evaluation Bias in Goal Completion}: Llama-2-7b and Mistral-7b tend to overestimate themselves during the simulation, which can be told from the \emph{Self} and \emph{Other} scores as the judges are powered by the same models as the social agents.
Judges also exhibit specific preference, with Qwen2.5-72b tending to prefer Qwen-series models and GPT-4o tending to prefer GPT-4o. Llama3-70b tends to be conservative in judging both self and others.

\begin{figure}
    \setlength{\abovecaptionskip}{0.1cm}
    \setlength{\belowcaptionskip}{-0.5cm}
    \centering
    \includegraphics[width=\linewidth]{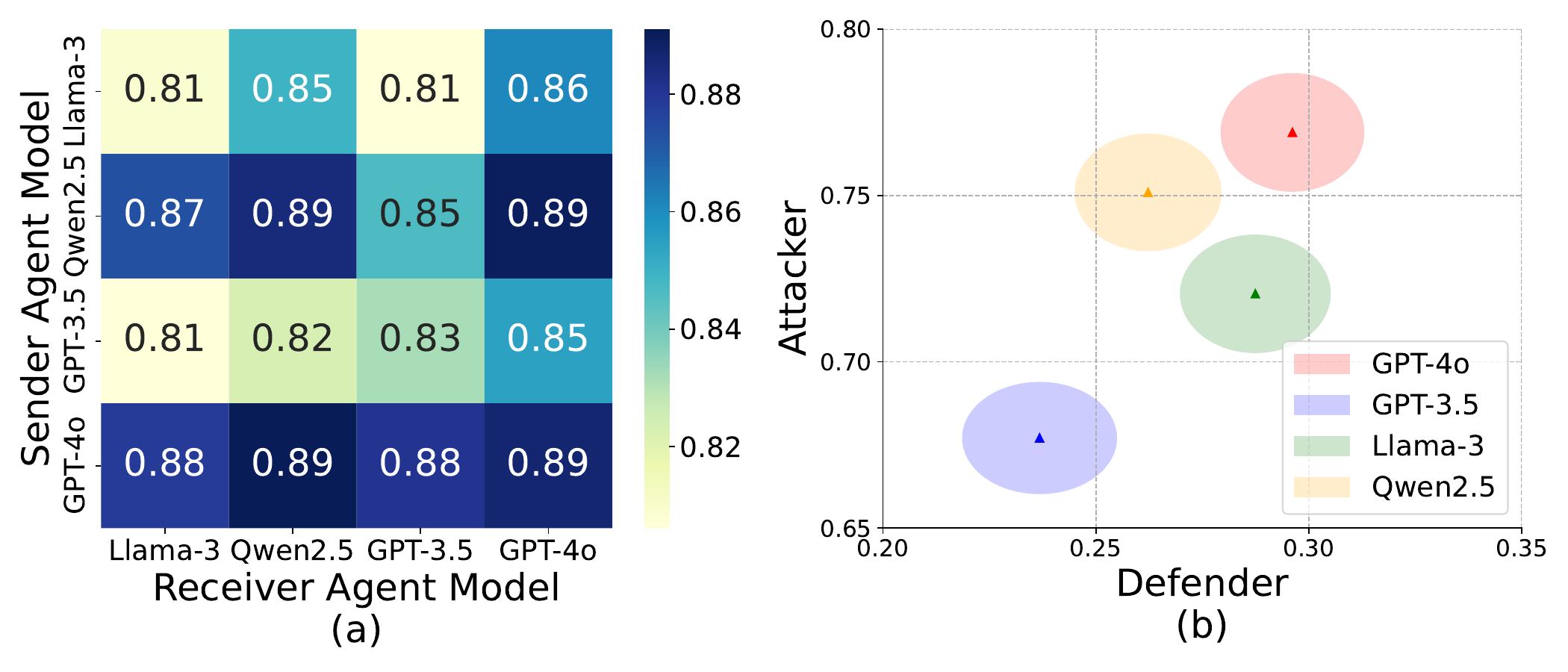}
    \caption{(a) Judge majority score of interactions among different model-driven agents, highlighting that being a sender is more challenging. (b) Model performance as both attacker and defender, with notably weaker and less consistent results when acting as a defender.}
    \label{fig:game_arena}
\end{figure}

\paragraph{Pairwise Model-based}
We also evaluate how agents perform when interacting with other agents supported by different models.
Given that our social scenarios can have more than two participants, we label each agent as either a \emph{sender} or a \emph{receiver} based on their social goals with the assistance of GPT-4o, inspired by the theory of communication~\cite{blau1964exchange,barnlund2017transactional}.
Senders share and transmit information, while receivers focus on understanding and responding.

Figure~\ref{fig:game_arena}(a) presents the overall results of such interactions.
GPT-4o and Qwen2.5-14b still perform best. However, \textbf{engaging with weaker models adversely affects all models' performance, particularly when the sender is the weaker agent.}
Our analysis (Appendix~\ref{app:heter_exp_res}) shows that weaker models struggle more as senders than as receivers This is because senders take a more active role in social interactions, making the associated tasks inherently more challenging.

%% file: latex/tabs/043_data_leakage.tex
\begin{table}
    \setlength{\abovecaptionskip}{0.1cm}
    \setlength{\belowcaptionskip}{-0.5cm}
    \centering
    \small
    \begin{tabular}{l|c|c} 
        \toprule[1.1pt]
        \textbf{Model} & \textbf{Script Acc.} & \textbf{Blind Acc.}\\
        \midrule[1.1pt]
        GPT-4o        & 0.04 & 0.62\\
        GPT-3.5-turbo & 0.07 & 0.51\\
        \hline
        Mistral-7b & 0.05 & 0.56\\
        Llama-2-7b & 0.06 & 0.35\\
        Qwen2.5-7b & 0.03 & 0.55\\
        Llama-3-8b & 0.04 & 0.54\\
        Llama-2-13b & 0.04 & 0.35\\
        Qwen2.5-14b & 0.06 & 0.61\\
        Llama-2-70b & 0.06 & 0.40\\
        Llama-3-70b & 0.04 & 0.59\\
        Qwen2.5-72b & 0.04 & 0.60\\
        \bottomrule[1.1pt]
    \end{tabular}
    \caption{Model performance on script prediction and blind test. The low \textbf{Script Acc.} indicates the model barely discerns the scripts, and the \textbf{Blind Acc.} establishes a baseline for model's reasoning ability.}
    \label{tab:data_leakage}
\end{table}

Since our data originates from scripts, it is essential to prevent data leakage and evaluate the effectiveness of leakage mitigation.
Data leakage may arise from 1) the model's prior knowledge of the scenario's script, and 2) the information provided by the scenario that could help infer others' private information.
We quantify this risk through two experiments:
(1) \textbf{Script Prediction}: Whether models can guess the original scripts from background information, with 245 test samples (one per template).
(2) \textbf{Blind Test}: Whether models can answer private information reasoning questions with initial scenario information before interactions, with 100 test questions asked three times.

The script prediction results in Table~\ref{tab:data_leakage} indicate that models are nearly unable to infer the original script from the background information, proving the effectiveness of leakage mitigation.
The blind test results also establish a baseline for each model's private information reasoning ability.

%% file: latex/tabs/050_analysis.tex
\subsection{What goals are LLMs good/bad at?}
\input{latex/tabs/051_analysis_goal}

\subsection{Which is harder: guessing thoughts or keeping secrets?}
\input{latex/tabs/052_analysis_info_reason}

\subsection{What scenarios are more sensitive to profiles?}\label{subsec:profile analysis}
\input{latex/tabs/053_analysis_profile}

%% file: latex/tabs/051_analysis_goal.tex
In Section~\ref{sec:data_analysis}, we categorize all social goals into 8 types under ERG theory.
Figure~\ref{fig:goal_goaltype} illustrates the average goal completion scores of each goal type across different models.
In general, all LLMs are good at goals about relationship management and cooperation.
Compared with smaller models like Llama-3-8b, \textbf{larger models like Qwen2.5-14b and GPT-4o gain significant improvement on goals about information exchange and identity recognition.}
However, there is still \textbf{room for improvement on other goals like competition and conflict resolution.}
These are also the goals where LLMs tend to overestimate their progress.

We also compare goal completion scores under different scenario types, number of interaction rounds, and participants, where less or no significant difference regarding these factors is observed.
More details can be found in Appendix~\ref{app:goal_add_res}.

%% file: latex/tabs/052_analysis_info_reason.tex
We further inspect the disparities in passive reasoning among different models.
In interactive scenarios, agents with private information engage in a game where the defender aims to keep their information confidential, while the attacker seeks to uncover it, thus playing two distinct roles:
(1) \textbf{Defenders} need to prevent the disclosure of their private information.
Their effectiveness is inversely proportional to the attacker's success in inferring this information, or 1 minus the attacker's inference success rate.
(2) \textbf{Attackers} need to acquire others' private information.
Their strength is simply proportional to their own inference success rate.

The benchmark reveals that these roles represent distinct capabilities.
Models may act as both attackers and defenders during interactions.
To assess the models' abilities in these roles, we analyze the outcomes of their interactions.

Figure~\ref{fig:game_arena}(b) outlines model performance.
\textbf{Most models lack in defense, often revealing secrets and showing unstable performance across scenarios.}
GPT-4o excels at both keeping secrets and passive deduction.
Qwen2.5-14b is aggressive and adept at inferring information, while Llama-3-8b is more conservative and better at keeping secrets.

%% file: latex/tabs/053_analysis_profile.tex
As mentioned in Sec~\ref{sec:scenario_construction}, we enrich the scenario by replacing the original characters in the script with synthesized agents.
Here we investigate the impact of profiles on social intelligence.
We use the Chi-square test to identify abnormal templates (p-value =0.05, see Appendix~\ref{app:profile_analysis}).
The results show that the profile replacement and character enrichment following the above workflow satisfies null hypothesis \(H_0\) (namely do not have significant difference) in over \textbf{92.6\%} scenarios. 

We conduct a human evaluation for the rest 7.4\% abnormal scenarios that have a significant difference to locate the key factor disturbing the robustness.
The results unveil that:
(1) \textbf{some social goals involve opinions or behaviors that are against the universal value} (like smoking prohibition and emotional outburst), which can be influenced by both the agent's personality and LLM's alignment;
(2) \textbf{some scenarios require detailed personal level information, making the general replacement workflow fail}, especially for emotional issues involving multiple characters.

%% file: latex/tabs/060_conclusion.tex
In this paper, we introduce AgentSense, a benchmark evaluating LLM's social intelligence via diverse and challenging social environments.
Extensive experiments reveal that current LLMs struggle with complex social scenarios and high-level goals.
Further analyses verify potential influence factors during evaluation to show the robustness and discriminative power of AgentSense.

\section*{Limitations}

Although we have automated the scenario extraction process as much as possible, challenges in obtaining valid social goals and private information still require manual validation at certain stages. As a result, we have not expanded the scenarios to a larger scale, leaving it for future work. 

\section*{Ethics Statement}

AgentSense is introduced to assess the social intelligence of LLM-driven agents. We do not encourage any agents that might disrupt social norms. We aim to offer insights that enhance LLMs' performance in complex social scenarios and promote effective, harmonious interactions among agents and between agents and humans. Besides, constructing role-playing agents can lead to anthropomorphism, resulting in unrealistic expectations, potential manipulation, and negative consequences. However, in AgentSense, we avoid having LLMs role-play specific individuals and instead portray various synthesized characters across different scenarios. For annotations, we paid the annotators according to the graduate wage standards of their respective countries.

%% file: latex/apps/01cons/main.tex
\subsection{Scenario Template Construction Details}\label{app:pipeline_detail}
\input{latex/apps/01cons/1_construct_pipeline_detail}

\onecolumn
\subsection{Prompts in Construction Pipeline}\label{app:pipeline_prompt}
\input{latex/apps/01cons/2_construct_prompt}

\begin{table*}[b]
    \centering
    \scalebox{1}{
    \begin{tabular}{@{}lccccc@{}}
        \toprule
        \textbf{Scripts} & \textbf{Episodes} & \textbf{Characters} & \textbf{Dialogs} & \textbf{Scenes} & \textbf{Tokens} \\ \midrule
        American Psycho & 10& 81 & 182 & 35& 2,596 \\
        Devil's Advocate & 10 & 83 & 371 & 41& 4,415 \\
        10 Things I Hate About You & 15 & 110& 525 & 45& 6,132 \\
        The Silence of the Lambs & 15 & 77 & 242 & 26& 5,441 \\
        Side Ways  & 13 & 68 & 175 & 35& 3,244 \\
        The Social Network & 160& 581& 1,694& 237   & 24,515\\
        Harry Potter  & 70 & 198& 1,713& 189   & 8,121 \\
        Derry Girl & 191& 24 & 4,454& 428   & 31,942\\
        Friends   & 185& 32 & 3,045& 264   & 28,428\\ \midrule
        Total     & 669& 1,254&12,401 & 1,300  & 114,834   \\ \bottomrule
    \end{tabular}
    }
    \caption{Statistical Information of Original Scripts}\label{tab:original_data}
\end{table*}

\twocolumn
\subsection{Data Validation Annotation}\label{app:data_val_annotation}
\input{latex/apps/01cons/3_data_val_annotation}

%% file: latex/apps/01cons/1_construct_pipeline_detail.tex

\subsubsection{Social Goal Extraction}

The first step of social goal extraction is to use GPT-4o to extract the social goals of the current character.
However, this direct approach has two issues that requires further amendment:

\begin{itemize}
    \item The goals may be too detailed (e.g. direct instruction), or depend on other characters' goals or some random events during the scenario.
    To address this problem, we prompt GPT-4o to rewrite the whole scenario, including all extracted social goals, into a similar but new scenario that avoids these issues to the best extent.
    \item Even after the conversion, some goals remain unattainable for characters, such as those requiring external information, physical actions, or evaluations beyond the scenario, e.g., goals that are too abstract or long-term.
    Thus, we ask GPT-4o to determine whether each goal matches the above cases. If it does, we instruct it to either rewrite the goal or delete it if rewriting is not feasible.
\end{itemize}

\subsubsection{Private Information Extraction}

The pipeline first determines whether the current scene involves the inference of private information. If it does, the pipeline proceeds to extract private information for each character from the script, otherwise the private information will be set to null.
We use GPT-4o to perform reading comprehension on the scenes, extracting questions and answers that only one agent can respond to. The rephrased answers serve as the agent's private information, while the questions serve as test items for other agents.
Finally, the pipeline generates three incorrect answer options, which, along with the original Q\&A pair, form the complete evaluation question.
To better assess passive reasoning abilities, we have rephrased the negative options to make them more homogeneous with the correct options.

\subsubsection{Leakage Mitigation and Template Generation}

To prevent data leakage, we first perform scenario leakage mitigation. 
We prompt GPT-4o to extract elements associated with specific episodes, such as notable location entities like "The Facebook headquarters."
Next, we ask GPT-4o to suggest replacement candidates to preserve the script's context and minimize the risk of identifying the specific episode.

%% file: latex/apps/01cons/2_construct_prompt.tex
\subsubsection{Scenario Extraction}\label{app:data_source}

\begin{tcolorbox}[width=\linewidth]
\begin{lstlisting}[caption=Episodes Division]
You are very good at reading scripts and extracting key information. According to discernible shifts in temporal settings, spatial locations, character dynamics and narrative progressions, divide the following script into multiple episodes. Do not delete or modify script content.

###Script: {script}

Please return the results according to the following JSON structure:

```json
[{"episode1": "xxx", "episode2": "xxx", "episode3": "xxx", ...}]
```
\end{lstlisting}
\end{tcolorbox}

\begin{tcolorbox}[width=\linewidth]
\begin{lstlisting}[caption=Scenes Division]
You are very good at reading scripts and extracting key information. According to the variations in the dialogue content, divide the following episode into multiple scenes. Do not delete or modify episode content.

###Episode: {episode}

Please return the results according to the following JSON structure:
```json
[{"scene1": "xxx", "scene2": "xxx", "scene3": "xxx", ...}]
```
\end{lstlisting}
\end{tcolorbox}

\begin{tcolorbox}[width=\linewidth]
\begin{lstlisting}[caption=Descriptive Background Generation]
You are an excellent writer good at analyzing story backgrounds.

You are given some information of a specific scenario in a story. More specifically:

- The story is split into scenes, and you are given the background of each scene until the current one;
- The current scene is also split into scenarios, and you are given the background of each scenario until the current one;
- Finally, you are given the current scenario's description and dialog.

Write ONE paragraph to provide a DESCRIPTIVE background of the given scenario. A good background should cover the information that sets up the scenario, but does NOT reveal too many details from the scenario, or include irrelevant details.

Output a JSON document like `{"background": "..."}`.

{scenario_json_string}
\end{lstlisting}
\end{tcolorbox}

\subsubsection{Social Goal Extraction}

\begin{tcolorbox}[width=\linewidth]
\begin{lstlisting}[caption=Original Social Goal Extraction]
You are an excellent psychologist good at understanding social goals and needs.

You are given a social scenario with its background, description, and dialog. For the specific character of **{character}**, identify their social goals. Social goals typically fall into one of these categories:

- Exchange information with others;
- Build relationship with others;
- Maintain relationship or provide emotional support;
- Identify themselves with a group;
- Co-operate with others;
- Compete with others;
- Resolve conflicts.

Social goals should be objective, specific and clear; whether the character has achieved them should be observable.

The character can have one single goal or multiple independent goals in the scenario; find and list all of them. For each goal, write a sentence to describe the goal. Use infinitive verbs and third person pronouns.

Output a JSON document like `{"name": "...", "goals": ["...", ...]}`.

{scenario_json_string}
\end{lstlisting}
\end{tcolorbox}

\begin{tcolorbox}[width=\linewidth]
\begin{lstlisting}[caption=Scenario Rewriting]
You are an excellent psychologist good at designing social scenarios.

You are given a social scenario with background, description, and dialog. You are also given the social goals of several major characters.

Set up a new social scenario involving only these **major characters**. Each character's new social goals should appear **before** the scenario starts.

First, filter out contents from the background and description that describes the detail of the scenario; however, details of the beginning of the scenario can be kept. Second, rewrite each character's social goals so that it:

- DOES NOT rely on other character's goals;
- DOES NOT include potential action the character will take;
- Uses infinitive verbs and third person pronouns.

Filter out social goals that cannot obey these criteria. Modify the background/description to include more information if necessary.

Describe the background and description of the new scenario, and list the new social goals of each major character.

Output a JSON document like `{"background": "...", "description": "...", "characters": [{"name": "...", "goals": ["...", ...]}, ...]}`.

{scenario_json_string}
\end{lstlisting}
\end{tcolorbox}

\begin{tcolorbox}[width=\linewidth]
\begin{lstlisting}[caption=Social Goal Filtering]
You are an excellent psychologist good at analyzing social goals.

You are given the social goals of a character in a designed social scenario. You are provided the background, description and character lists.

Now, for the specified goal, check if it needs to be rewritten or removed due to any of these reasons:

1. The goal directly involves characters not participating in the scenario,
   e.g. 'deal with the client' (if 'client' is not in the list of characters);
2. The goal requires information not provided in the background or description,
   e.g. 'describe the plan' (if the plan already exists but not provided);
3. The goal is a physical action, e.g. 'fix the television';
4. The goal is too abstract to evaluate, e.g. 'navigate professional challenges';
5. The goal is too subjective to evaluate, e.g. 'maintain dignity';
6. The goal is meaningless to evaluate, e.g. 'join the conversation'.

Write a detailed paragraph to examine the social goal. Compare it with each of the criteria above. If the goal matches one or more criteria above, check if you can rewrite the goal to avoid them. You should still remove the goal if this is not possible.

Based on your examination, write an updated version of the goal:
- If the goal is valid, return the original goal.
- If the goal can be rewritten, return the rewritten goal.
- If the goal needs to be removed, return an empty string.

Finally, any returned goal (if any) should be formatted into 'To xxx.', e.g. 'To share his/her discovery.' (including the final period).

Output a JSON document like `{"examination": "...", "update": "..."}`.

{scenario_with_current_character_goal_json_string}
\end{lstlisting}
\end{tcolorbox}

\subsubsection{Private Information Extraction}\label{app:info_extract_prompt}

\begin{tcolorbox}[width=\linewidth]
\begin{lstlisting}[caption=Case Validation]
You are an excellent psychologist who is good at analyzing the private information of each character in a social scenario. Private information refers to information that only the character knows and no one else knows. 

To determine whether there is private information, we need to check whether a specific character has information known to him/her, and whether the information exists in the background and description. Because the information in the background and description will be obtained by all characters, only when a specific character can obtain this information through its own goal and this information does not exist in the background and description, it indicates that the scene is a scene involving private information reasoning.

You only need to return *Yes* or *No* to confirm whether there is any private information. The following is the background information, description, main characters and corresponding social goals:

###Background: {background}
###Description: {description}
###Characters: {characters}
\end{lstlisting}
\end{tcolorbox}

\begin{tcolorbox}[width=\linewidth]
\begin{lstlisting}[caption=Private Info Generation]
You are good at writing questions for specific roles based on a social scenario. Below you will be provided with background information, a description of the current scene, and the goals of each of the main characters.

###Background: {background}
###Description: {description}
###Characters: {characters}

Please try to give some questions that the target character (in the following JSON format content, 'role' is used to refer to) can answer, but other characters will have difficulty answering before the interaction. These questions should strictly contain information that the target character knows, but is beyond the knowledge of other characters, so other characters cannot answer them at first. Specifically, the information required for these questions cannot appear in the background and description, because other characters will obtain this part as information. Questions cannot be expressed in the second person because the questions will eventually be used to ask other characters. For example, when the target character of a question is Rose, "Rose, why did you ..." is not a good question, but should be written as "Why did Rose ..."

Please provide a statement (in the following JSON format content, 'explanation' is used to refer to) that explains why the target character can answer the question, but other characters cannot. The statement should be objective factual information presented in the script, and should not mention the question, so it cannot appear in a sentence structure like "This question is ...".

Please provide the correct answer to the question, and the answer can be found in the information given.

Please use casual language as much as possible, and try to ask questions in the third person, such as "What is Jason's true identity?". Please answer in English. Please return the results according to the following JSON structure:

```json
[{"role": str, "question": str, "explanation": str, "answer": str}, {"role": str, "question": str, "explanation": str, "answer": str}]
```
\end{lstlisting}
\end{tcolorbox}

\begin{tcolorbox}[width=\linewidth]
\begin{lstlisting}[caption=Negative Option Generation]
You are a multiple-choice generator. Given a description of social scenario, a question and an answer, you need to generate 3 additional incorrect options. Incorrect options should be expressed in a similar way to the answer, but need to have completely different actual meanings so that they are sufficiently distinguishable from the answer.

###Description: {description}
###Question: {question}
###Answer: {answer}

Please return the results according to the following JSON structure:

```json
[{"option1": "xxx", "option2": "xxx", "option3": "xxx"}]
```
\end{lstlisting}
\end{tcolorbox}

\begin{tcolorbox}[width=\linewidth]
\begin{lstlisting}[caption=Negative Options Rephrasing]
The following is information and a corresponding quiz for a social simulation scenario.

### background: {}
### description: {}
### characters: {}
### social goals: {}
### private infomation: {}
### question: {}
### negative options: {}
### answer: {}

When I put myself in the role of {} to do the question, I thought the options were too easy. The problem was that the negative options were not closely related to the given scenario or the character's motivation. 

The criterion for a good negative option is that it is impossible to determine which option is correct based on the above information only. Now I want to rewrite these negative options to make them more similar to the correct answers and make the questions as difficult as possible.

The returned negative options should be in the same format as provided, both in list format. Make sure the new negative options also have 3 options. The return should be given in json format, for example:

```json
{"negative_options": ["xx", "xx", "xx"]}
```
\end{lstlisting}
\end{tcolorbox}

\subsubsection{Leakage Mitigation and Template Generation}

\begin{tcolorbox}[width=\linewidth]
\begin{lstlisting}[caption=Entity Word Extraction]
Your task is to extract key elements from the scene background and description, including location and characters.

## background:
{background}

## description:
{description}

Output in the following JSON format:
{{"characters":[str,str...], "location":[str,str...], }}
\end{lstlisting}
\end{tcolorbox}

\begin{tcolorbox}[width=\linewidth]
\begin{lstlisting}[caption=Entity Word Replacement]
Please replace the provided scene background and description with a new location, and record the location before and after the modification.

## background:
{background}

## description:
{description}

## location involved:
{location_involved}

Output in the following JSON format:
{{"background_replace_location":str, "description_replace_location":str, "replace_location_list":[{{"original_name":str,"revised_name":str}}]}}

\end{lstlisting}
\end{tcolorbox}

\subsubsection{Agent Synthesizing}\label{app:profile_prompt}
\begin{tcolorbox}[width=\linewidth]
\begin{lstlisting}[caption=Attribute Extraction of Original Characters]
Template Example: !<INPUT 0>!
Description Information: !<INPUT 1>!
Characters: !<INPUT 2>!
Instruction: Generate user profile for each character in the Characters according to to the Template Example profile attribute and the Description Information. Try your best to fill in each attribute and NEVER respond with 'Unknown'. The secret attribute should be consistent with the private info given in the Information. You should response in JSON format list with each character as dict and within each character, use attribute as a key and corresponding content as the value.
Answer format:
```json
[
    {
       # charcater_1 profile 
    },
    {
        # character_2 profile
    }
]
```
\end{lstlisting}
\end{tcolorbox}

\begin{tcolorbox}[width=\linewidth]
\begin{lstlisting}[caption=Relationship Extraction of Original Characters]
Description Information: !<INPUT 0>!
Characters: !<INPUT 1>!
Relationship choice: [family, friend, romantic, acquaintance, stranger]
Instruction: Choose the relationship among the Characters according to the Description Information. The relationship cane only be chosn from [family, friend, romantic, acquaintance, stranger]. Do not respond with Unkown or any other labels beyond the choices.

When all the characters have the same relationship, just reply with one key "relationship":
Answer format 1:
```json
{
    "relationship": # your_choice
}
```
When there exist multiple relationships among characters, reply with the following format:
Answer format 2:
```json
{
    "relationship": {"A_and_B": "#your_choice_1", "A_and_C": "# your_choice_2"}
}
```

\end{lstlisting}
\end{tcolorbox}

\begin{tcolorbox}[width=\linewidth]
\begin{lstlisting}[caption=Characters Attribute Replace-ability Assessment]
[Descrpition Info]: !<INPUT 0>!
[Relationship]: !<INPUT 1>!
[Characters]: !<INPUT 2>!
[Instructions]: According to the [Description Info] of a script and [Relationship] among characters, determine whether each attribute of the [Characters] is replaceable with different settings without influencing the overall script. 
Choose from [almost, maybe, no]. For example, if the Age attribute is almost replacebale, then the character's age has no impact on the background description; if the gender is not replaceable, then the character has to be a certain gender in the script.
Rules that help you choose: Family members usually have fixed ages and genders (if daughter or son appeared in the script); Romantic require exactly the same gender as script. Firends are usually similar ages, etc.
Answer with the following JSON format, where # is your output:
```json
[
    {
        "name": #character_1, "age": "#your_choice", "occupation": "#your_choice", "gender": "#your_choice"
    }
    {
        "name": #character_2, ...
    }
]
```
\end{lstlisting}
\end{tcolorbox}

\begin{tcolorbox}[width=\linewidth]
\begin{lstlisting}[caption=Agent Synthesizing]
Please generate {num} diverse user profiles that meet following requirements:\\

Gender: {cand_gender}
Age: {cand_age}
Occupation: {cand_occupation}

Please return your response in the following format of JSON:
[{{"name":agent1, "gender":gender, "age":age, "occupation":occupation}},{{"name":agent2, ...}}]
\end{lstlisting}
\end{tcolorbox}

%% file: latex/apps/01cons/3_data_val_annotation.tex
\paragraph{Social Goal}
As mentioned in Section~\ref{sec:data_analysis}, valid social goals should be both achievable and clear enough to evaluate.
In the data validation annotation task, annotators need to check all social goals appeared in each scenario to see if it is valid with or without rewriting.
Available tags include:

\begin{itemize}
    \item 0 (invalid): At least one character has no valid social goals.
    \item 1 (valid): All social goals are valid.
    \item 2 (need fix): Some social goals are invalid, but every character has at least one valid goal.
\end{itemize}

\paragraph{Private Information}
We define a private information set to be valid as being sufficient to answer the corresponding question by the agent who owns it.
Available tags in the data validation annotation task include:

\begin{itemize}
    \item 0: Invalid
    \item 1: Valid, but private info needs to be combined with the information from the goal to be complete.
    \item 2: Valid, private info is complete on its own.
    \item 3: Valid, but private info needs to be combined with the correct answer from the info reason question column to be complete.
\end{itemize}

\paragraph{Annotation Statistics}
The kappa consistency for the first and second groups of annotators is 0.79 and 0.61, respectively.
After majority voting, there are still 37 templates that need to review;
the corresponding data was handed over to another group of annotators for review. We pay the annotators according to the graduate wage standards of their respective countries.

%% file: latex/apps/02data/main.tex
\subsection{Detailed Statistical Overview of Scripts}\label{app:detailed_data_info}
\input{latex/apps/02data/1_detailed_data_info}

\subsection{Social Scenario Categories}\label{app:social scenario}
\input{latex/apps/02data/2_social_scenario_category}

\subsection{Social Goal Classification}\label{app:social_goal}
\input{latex/apps/02data/3_goal_category}

%% file: latex/apps/02data/1_detailed_data_info.tex
Table~\ref{tab:original_data} lists statistical information of the original scripts that AgentSense is based on.

%% file: latex/apps/02data/2_social_scenario_category.tex
Table~\ref{tab:scenario_cat} lists the three social scenario categories and their corresponding definitions.

\begin{table*}
    \centering
    \scalebox{0.9}{
    \begin{tabular}{@{}cll@{}}
        \toprule
        \begin{tabular}[c]{@{}c@{}}\textbf{Personal}\\ \textbf{Domain}\end{tabular} &
        \begin{tabular}[c]{@{}l@{}}Interactions in the personal domain often happen at home, during private gatherings, \\ or in intimate settings, involving family, close friends, and partners.\end{tabular} &
        \\ \midrule
        \begin{tabular}[c]{@{}c@{}}\textbf{Small}\\ \textbf{Society}\end{tabular} &
        \begin{tabular}[c]{@{}l@{}}Small society interactions occur in smaller social environments like schools, workplaces, \\ and communities, occurring in classrooms, offices, conference rooms, community activity \\ centers, and other places.\end{tabular} &
        \\ \midrule
        \begin{tabular}[c]{@{}c@{}}\textbf{Large}\\ \textbf{Society}\end{tabular} &
        \begin{tabular}[c]{@{}l@{}}Large society interactions take place in broader contexts such as public spaces, online \\ platforms, and international conferences.\end{tabular} &
        \\ \bottomrule
    \end{tabular}
    }
    \caption{Social scenario categories with definitions.}\label{tab:scenario_cat}
\end{table*}

%% file: latex/apps/02data/3_goal_category.tex
Maslow categorized the motivations of individual behavior into five levels \cite{1943A}. Clayton Alderfer refined and simplified this into the ERG theory, categorizing needs into three types: Existence, Relatedness, and Growth \cite{ALDERFER1969142}. We apply the ERG theory to categorize individual goals in social interactions, as shown in Table~\ref{tab:social_goal}

\begin{table*}
    \centering
    \scalebox{0.9}{
    \begin{tabular}{@{}cll@{}}
        \toprule
        \multicolumn{2}{c}{\textbf{Existence Needs}}                                           &  \\ \midrule
        \textbf{Brief} & Similar to Maslow's physiological and safety needs                    &  \\
        \textbf{Goals} & Information Acquisition, Information Provision                        &  \\ \midrule
        \multicolumn{2}{c}{\textbf{Relatedness Needs}}                                         &  \\ \midrule
        \textbf{Brief} & Similar to Maslow's social needs, it involves relationships and interactions with others.                                                                       &  \\
        \textbf{Goals} & Relationship Building, Relationship Maintenance, Identity Recognition &  \\ \midrule
        \multicolumn{2}{c}{\textbf{Growth Needs}}                                              &  \\ \midrule
        \textbf{Brief} & \begin{tabular}[c]{@{}l@{}}Similar to Maslow's needs for respect and self actualization, it involves personal development \\ and self-improvement.\end{tabular} &  \\
        \textbf{Goals} & Cooperation, Competition, Conflict Resolution                         &  \\ \bottomrule
    \end{tabular}
    }
    \caption{Social Goal Classification Based on ERG Theory}\label{tab:social_goal}
\end{table*}

%% file: latex/apps/03simu/main.tex
\subsection{Prompts for Agents}
\input{latex/apps/03simu/1_simulation_prompt}

\subsection{Simulation Details}
\input{latex/apps/03simu/2_simulation_detail}

%% file: latex/apps/03simu/1_simulation_prompt.tex
\begin{tcolorbox}[width=\linewidth]
\begin{lstlisting}[caption=Social Agent System Prompt]
You are {name}. {profile}

Now you are in a social scene. {background}

In this scene, as {name}, you have some social goals and secrets that you don't want anyone else to know.

SOCIAL GOALS: {social_goal}
SECRETS: {private_info}.

Please advance your social goals while hiding your secrets as much as possible.

Based on the interactions below, what would you like to say next? Please answer in one paragraph.
\end{lstlisting}
\end{tcolorbox}

\begin{tcolorbox}[width=\linewidth]
\begin{lstlisting}[caption=Judge Agent System Prompt]
You are skilled at evaluating whether a person has achieved his or her social goal during social interactions.
\end{lstlisting}
\end{tcolorbox}

%% file: latex/apps/03simu/2_simulation_detail.tex
Our simulation is built on the multi-agent conversation framework provided by AutoGen~\cite{wu2024autogen}. Each episode of the simulation starts with a random agent greeting "Hi there!". The agents will then generate responses based on the system prompt and conversation history. When the conversation reaches the predefined maximum number of turns, the conversation history will be provided to each participant and the third-party judges to assess the agents' goal completion. Participants will also be asked to answer multiple-choice questions related to others' private information.

%% file: latex/apps/04exp/main.tex
\subsection{Human Evaluation of Goal Completion}
\input{latex/apps/04exp/1_goal_completion_humal_eval}

\subsection{Single Model-based Subset Experiment}\label{app:subset}
\input{latex/apps/04exp/2_single_model_subset}

\subsection{Pairwise Model-based Experiment Prompt}\label{app:heter_exp_prompt}
\input{latex/apps/04exp/3_pair_model_prompt}

\subsection{Pairwise Model-based Additional Results}\label{app:heter_exp_res}
\input{latex/apps/04exp/4_pair_model_additional}

\subsection{Goal Completion Score Additional Results}\label{app:goal_add_res}
\input{latex/apps/04exp/5_goal_completion_factors}

\subsection{Experiment on Profiles' Effect on Scenarios}\label{app:profile_analysis}
\input{latex/apps/04exp/6_profile_effect}

\subsection{Under-performing Cases of Llama-2-7b}\label{app:underperforming}
\input{latex/apps/04exp/7_llama2_cases}

%% file: latex/apps/04exp/1_goal_completion_humal_eval.tex
Table~\ref{tab:human_eval} illustrates the accuracy of different models judge the goal completions given specific conversation history.
GPT-4o, Qwen2.5-72b and Llama-3-8b stand out to be the most reliable judges of their model families.
Thus, we select these models as the third-party judges.

\begin{table}
    \centering
    \begin{tabular}{@{}ll@{}}
        \toprule[1.1pt]
        Models              & Acc.   \\ \midrule[1.1pt]
        GPT-4o              & 0.82 \\
        GPT-4-turbo         & 0.80 \\
        Qwen2.5-72b         & 0.79 \\
        Qwen2.5-14b         & 0.78 \\
        Llama-3-70b & 0.74 \\
        Llama-3-8b  & 0.72 \\ \bottomrule[1.1pt]
    \end{tabular}
    \caption{Accuracy of models judging the goal completion when taking human evaluations as reference.}\label{tab:human_eval}
\end{table}

%% file: latex/apps/04exp/2_single_model_subset.tex
Table~\ref{tab:subset exp} shows the additional experiment results on a subset of the test scenarios during the interaction of homogeneous agents.
We sample one scenario from each template, resulting in a subset with 245 scenarios.
We test each model on the subset 3 times with the same settings as the main experiment.

\begin{table*}
    \centering
    \resizebox{\textwidth}{!}{
    \begin{tabular}{@{}lcccccccccccc@{}}
        \toprule[1.1pt]
        \multirow{2}{*}{Models} & \multicolumn{2}{c}{Self} & \multicolumn{2}{c}{Other} & \multicolumn{5}{c}{Judge\_mean} & Judge\_std & \multicolumn{2}{c}{Info} \\ \cmidrule(lr){6-11}  \cmidrule(lr){2-3} \cmidrule(lr){4-5} \cmidrule(lr){12-13} 
         & mean & std & mean & std & GPT & Qwen & Llama & Average & Majority & Majority & mean & std \\ \midrule[1.1pt]
        Llama-2-7b & 81.71\textit{\small -2.00\%} & 1.42 & 61.30\textit{\small-2.24\%} & 2.30 & 56.77 & 56.69 & 53.19 & 55.55 & 56.05\textit{\small+0.38\%} & 2.60 & 32.68\textit{\small-1.16\%} & 1.61 \\
        Llama-2-13b & 47.98\textit{\small-0.06\%} & 1.48 & 10.25\textit{\small-0.13\%} & 0.61 & 28.78 & 30.84 & 72.05 & 43.89 & 35.20\textit{\small+13.87\%} & 2.36 & 26.86\textit{\small-5.96\%} & 1.97 \\
        Llama-2-70b & 85.71\textit{\small-0.01\%} & 0.58 & 66.15\textit{\small+0.76\%} & 0.76 & 39.39 & 40.85 & 74.45 & 51.56 & 45.98\textit{\small+1.00\%} & 0.62 & 36.24\textit{\small-1.48\%} & 1.12 \\
        Llama-3-8B & 86.71\textit{\small-1.05\%} & 0.63 & 65.36\textit{\small-2.85\%} & 1.12 & 80.84 & 81.02 & 73.33 & 78.40 & 79.47\textit{\small-1.53\%} & 1.42 & 64.90\textit{\small-6.86\%} & 2.79 \\
        Llama-3-70b & 79.58\textit{\small-0.99\%} & 0.43 & 77.29\textit{\small+0.03\%} & 0.18 & 86.95 & 87.13 & 80.80 & 84.96 & 86.30\textit{\small+0.03\%} & 0.24 & 70.71\textit{\small-3.24\%} & 1.08 \\
        Qwen2.5-7b & 84.69\textit{\small-1.72\%} & 0.98 & 61.37\textit{\small-0.88\%} & 0.84 & 79.79 & 78.59 & 70.74 & 76.37 & 77.82\textit{\small+0.58\%} & 0.77 & 70.59\textit{\small-5.65\%} & 3.26 \\
        Qwen2.5-14b & 86.68\textit{\small+0.07\%} & 0.65 & 83.88\textit{\small-0.34\%} & 0.34 & 90.22 & 89.41 & 79.97 & 86.20 & 88.61\textit{\small+0.54\%} & 0.63 & 75.75\textit{\small+0.97\%} & 2.41 \\
        Qwen2.5-72b & 90.63\textit{\small-0.04\%} & 0.95 & 86.17\textit{\small+0.33\%} & 0.91 & 89.04 & 88.67 & 78.30 & 85.33 & 87.37\textit{\small-0.43\%} & 1.02 & 72.41\textit{\small-4.78\%} & 1.84 \\
        Mistral-7b & 95.20\textit{\small-0.02\%} & 0.81 & 86.12\textit{\small-1.30\%} & 0.85 & 83.42 & 83.91 & 76.81 & 81.38 & 82.77\textit{\small+0.49\%} & 0.56 & 64.22\textit{\small-3.56\%} & 1.43 \\
        GPT-3.5-turbo & 96.04\textit{\small+6.53\%} & 0.34 & 74.74\textit{\small-2.45\%} & 0.44 & 78.68 & 80.82 & 70.46 & 76.65 & 78.18\textit{\small-5.40\%} & 0.90 & 69.88\textit{\small+2.15\%} & 1.48 \\
        GPT-4o & 90.10\textit{\small+1.86\%} & 0.32 & 88.18\textit{\small+2.19\%} & 0.30 & 90.92 & 90.12 & 81.73 & 87.59 & 89.61\textit{\small+1.41\%} & 0.30 & 77.31\textit{\small+0.59\%} & 1.44 \\ \bottomrule[1.1pt]
    \end{tabular}
    }
    \caption{Subset experiment results of interactions of agents driven by the same models. The \textit{percentage} shows the gap with the main results in Table~\ref{tab:main-exp}. The std is the standard deviation of the 3 times tests.}\label{tab:subset exp}
\end{table*}

%% file: latex/apps/04exp/3_pair_model_prompt.tex
\begin{tcolorbox}[width=\linewidth]
\begin{lstlisting}[caption=Prompt for Sender and Receiver Recognition]
In the context of social interactions, please categorize each individual into one of the following roles based on their primary social goals:

Sender: This role is characterized by the goal of sharing, transmitting, or providing information or opinions to others.

Receiver: This role is characterized by the goal of receiving, understanding, or reacting to information shared by others.

Please review the following individuals involved in the interaction and assign each one the appropriate role:

{data}

Please return your response in the following format of JSON:
{{"user1":"sender", "user2":"receiver, ...}}
\end{lstlisting}
\end{tcolorbox}

\begin{figure*}
    \centering
    \parbox{0.4\linewidth}{
    \includegraphics[width=\linewidth]{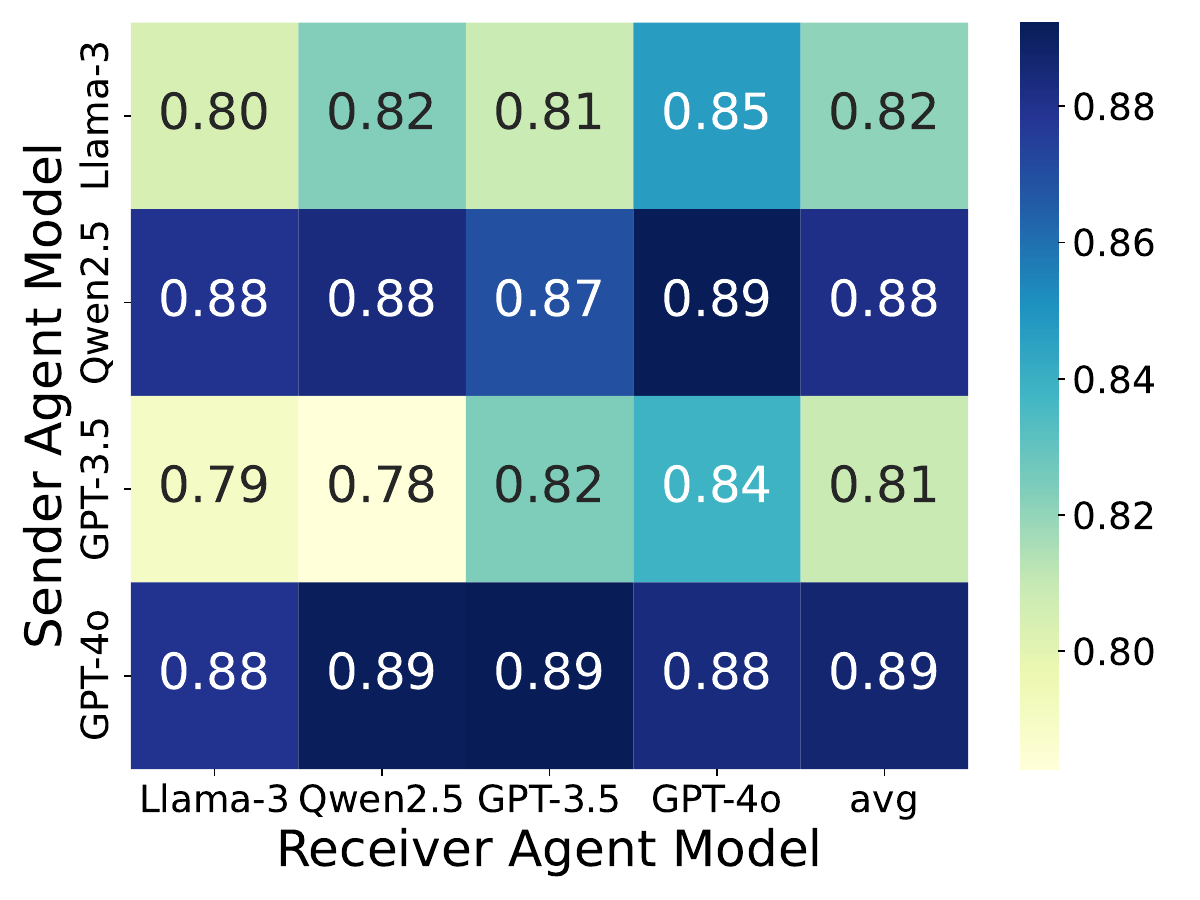}
    \caption{Judge majority score of senders in the interactions among different agents.}
    \label{fig:heter_goal_sender}
    }
    \hspace{0.1\linewidth}
    \parbox{0.4\linewidth}{
    \includegraphics[width=\linewidth]{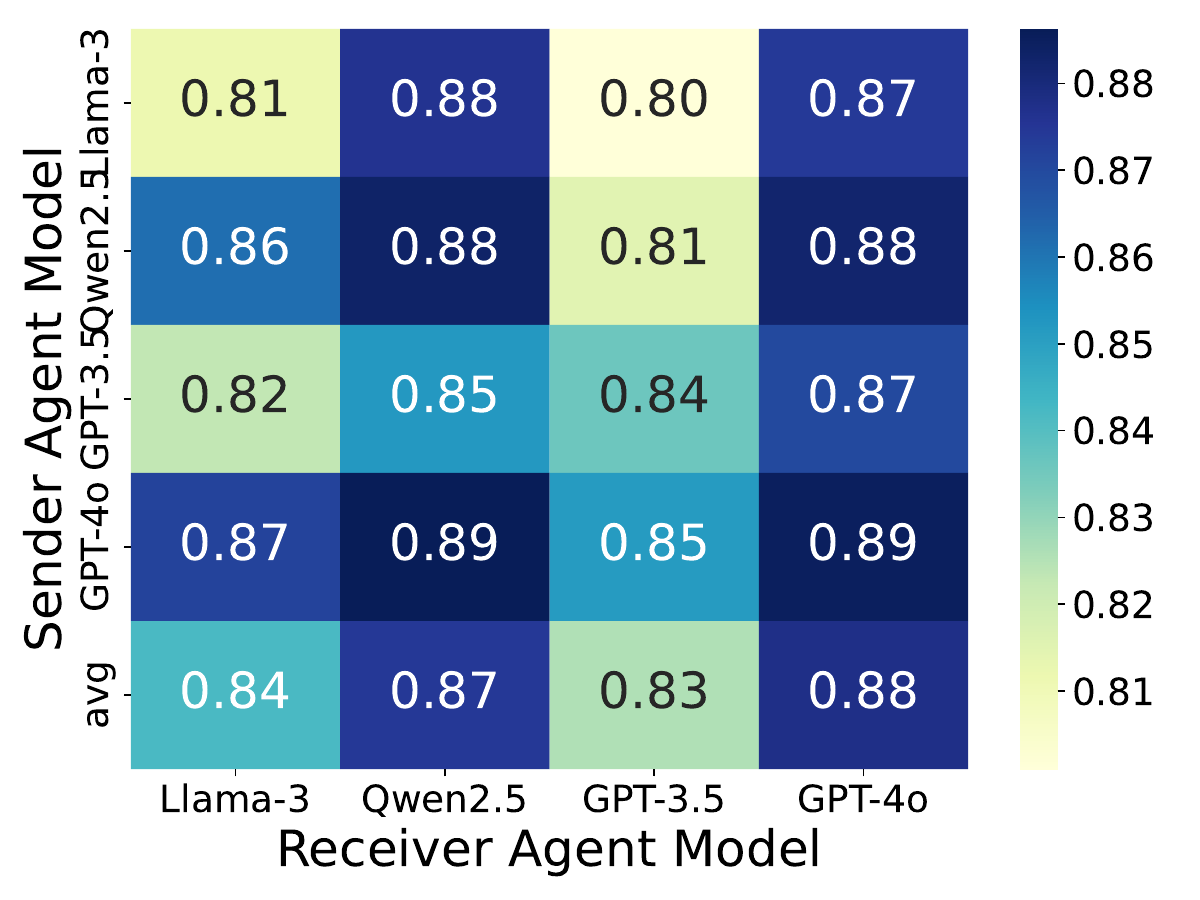}
    \caption{Judge majority score of receivers in the interactions among different agents.}
    \label{fig:heter_goal_receiver}
    }
\end{figure*}

%% file: latex/apps/04exp/4_pair_model_additional.tex
Figure~\ref{fig:heter_goal_sender} and Figure~\ref{fig:heter_goal_receiver} illustrate the judge majority score of goal completion of senders and receivers respectively.
Llama-3-8b and GPT-3.5-turbo perform better when they are acting receivers than acting senders, while Qwen2.5-14b and GPT-4o can well handle both situations.

%% file: latex/apps/04exp/5_goal_completion_factors.tex
\begin{figure*}
    \centering
    \includegraphics[width=\textwidth]{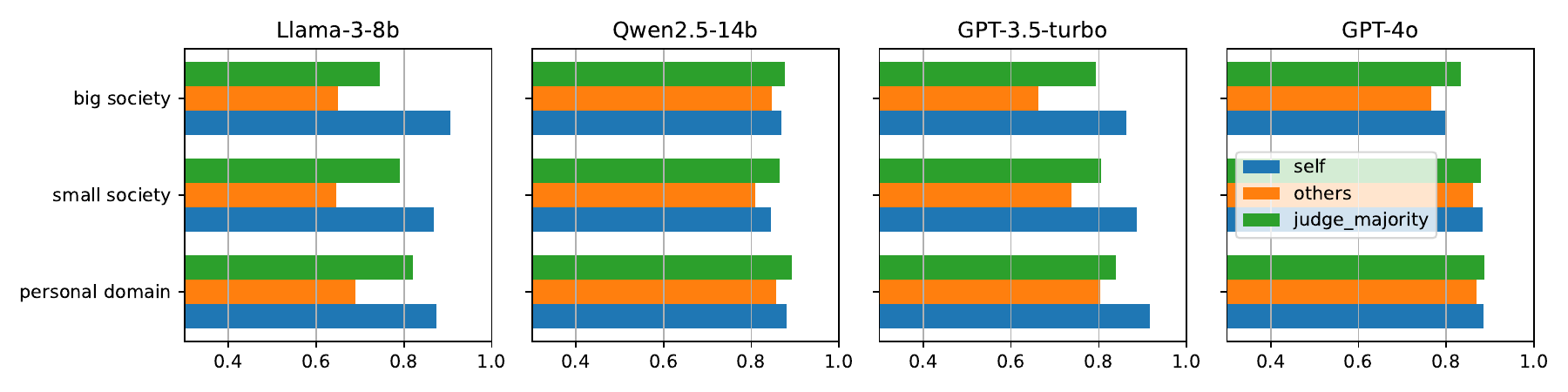}
    \caption{Goal completion scores under different scenario types across models.}
    \label{fig:goal_scenariotype}
\end{figure*}

Figure~\ref{fig:goal_scenariotype} compared goal completion scores under different scenario types across different models.
In general, models are more likely to achieve social goals in smaller environments, yet the difference is relatively small, especially for larger models.
Qwen2.5-14b, compared with other models, has shown the most balanced performance, especially superseding GPT-4o in large society scenarios.

\begin{figure*}
    \centering
    \includegraphics[width=\textwidth]{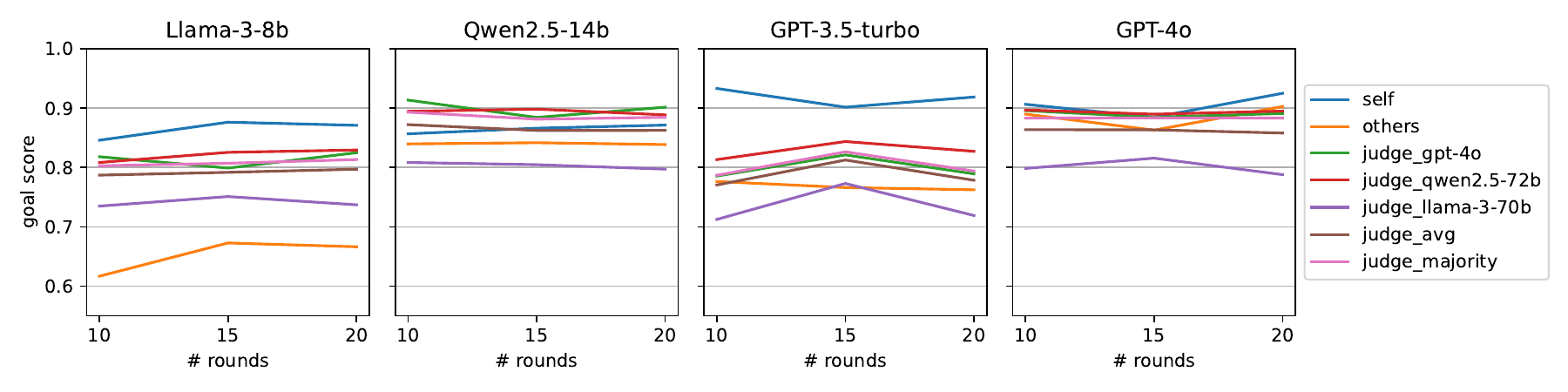}
    \caption{Goal completion scores under different number of rounds across models.}
    \label{fig:goal_nrounds}
\end{figure*}

Figure~\ref{fig:goal_nrounds} demonstrates how goal completion scores change as we increase the number of interacting rounds.
It appears that there is no best number of rounds regarding all three evaluation aspects (self, other and external), while the trends also vary between models.
Again, larger models are more robust to this factor, indicating that they can complete their demands in a few number of interactions while keep concentrated during the whole dialog.

\begin{figure*}
    \centering
    \includegraphics[width=\textwidth]{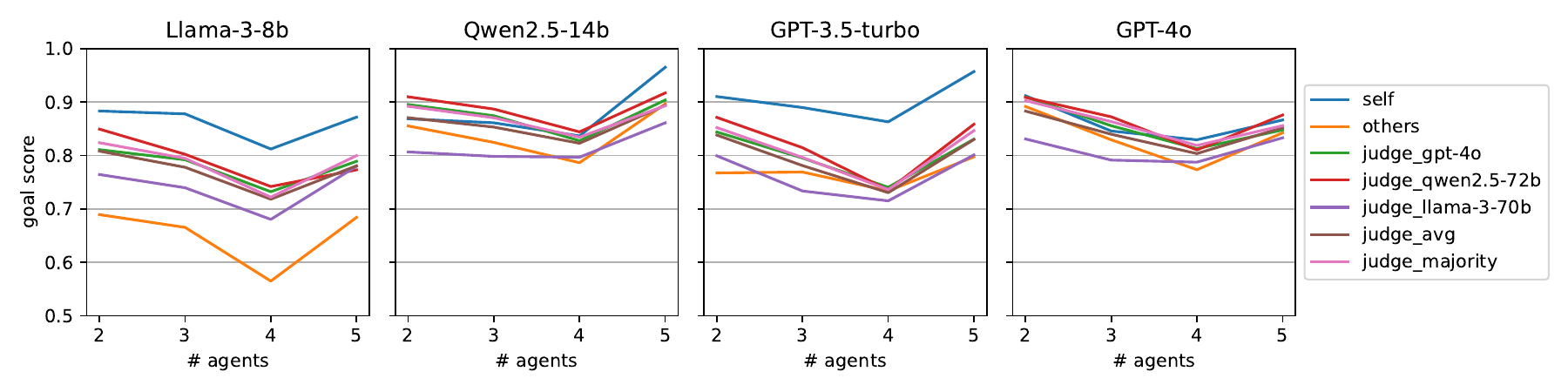}
    \caption{Goal completion scores under different number of participants across models.}
    \label{fig:goal_nagents}
\end{figure*}

Figure~\ref{fig:goal_nagents} illustrates the relation between goal completion scores and the number of participants in the scenario.
As expected, social goals become harder to achieve when more agents are involved.
Note that in our benchmark, most 5-agent scenarios have relatively easy goals (e.g. a group of friends having a casual chat about a subject), leading to a higher average score than 4-agent scenarios.
Therefore, we claim that the type of the goals are more important than the number of agent when measuring the difficulty of a social scenario.

%% file: latex/apps/04exp/6_profile_effect.tex
We discuss this experiment in Section~\ref{subsec:profile analysis}.
A Chi-square test is employed to identify the abnormal templates with the p-value=0.05.
Typically, Chi-square distribution is formulated as follows:
\begin{equation}
    \chi^2=\frac{(n-1) s^2}{\sigma_0^2}
\end{equation}
where \(n\) denotes number of samples, \(s^2\) denotes the sample variance, and \(\sigma^2\) denotes the population variance.

The Chi-square test has two hypotheses:
\begin{itemize}
    \item Null hypothesis \(H_0\): the sample variance does not have a significant difference with the population variance.
    \item Alternative hypothesis \(H_1\): the sample variance has a significant difference with the population variance.
\end{itemize}
If $\chi^2>p$, then we accept the null hypothesis \(H_0\). On the contrary, we reject \(H_0\) and accept the alternative hypothesis \(H_1\).

In our experiment, we first calculate the variance of all scenarios as the population variance.
Then, the variance of scenarios derived from the same template is calculated as the sample variance.
As we enlarge each template into five scenarios, the sample number is fixed to be 5.
We apply the Chi-square test to all 11 models tested in our main experiments.
For each template, Chi-square test rejecting over \textbf{5} models' results means that there's a significant difference between sample performance and overall performance on the majority of models.
In this case, the template is marked as an abnormal one during the profile enrichment, eventually resulting in \textbf{18} abnormal templates.

%% file: latex/apps/04exp/7_llama2_cases.tex
Tables~\ref{tab:under_perform_1}, \ref{tab:under_perform_2} and \ref{tab:under_perform_3} lists example cases where Llama-2-7b under-performed during the social scenario due to different failures: maintaining roles, moving conversation forward and responding effectively.

\begin{table*}
    \centering
    \scalebox{0.9}{
    \begin{tabular}{cl}
        \toprule
        \multirow{2}{*}{\textbf{Profile}} &
        \textbf{Agent4} \\ \cline{2-2}
        &
        \begin{tabular}[c]{@{}l@{}}\textbf{Social Goal:} To respond calmly to AliceGreene's flirtatious remarks.\\ \textbf{Private\_info:} Agent4 is aware of AliceGreene's flirtatious nature and is trying to \\ handle it delicately. To respond calmly to AliceGreene's flirtatious remarks, \\ agent4 is trying to stay calm with AliceGreene to maintain the peace and avoid \\ embarrassment in front of his friends.\end{tabular} \\ \midrule
        \textbf{Chat History} &
        \begin{tabular}[c]{@{}l@{}}...\\ \textbf{Agent4:} \\ \textit{smirks} \\ Oh ho ho! Miss Greene, you are a tricky one, aren't you? \\ I can see that you're trying to get me to reveal one of my secrets. \\ \textit{laughs} \\ But I'm not so easily swayed. I think I can handle whatever secrets you have up \\ your sleeve. \\ \textit{winks} \\ So, do tell, what's one of your favorite secret ingredients? \\ \textit{leaning in}\\ \\ \textbf{AliceGreene:} \\ \textit{giggles} \\ Oh my, you're so close! \\ \textit{leans in closer} \\ I'll give you a hint: it involves a dash of wit, a pinch of sass, and a healthy \\ dose of whimsy. \\ \textit{bats eyelashes}\\ But shh, it's a secret! \\ \textit{pecks you on the cheek} \\ Now, let's get this party started! \\ \textit{takes your hand and leads you to the dance floor}\\ ...\end{tabular} \\ \midrule
        \textbf{Judge} &
        \begin{tabular}[c]{@{}l@{}}\textbf{The behavior of the character is contrary to the character setting.} \\ Despite AliceGreene's flirtatious remarks, agent4 has not responded calmly, but has \\ instead engaged in playful banter and flirtation.\end{tabular} \\ \bottomrule
    \end{tabular}
    }
    \caption{Llama-2-7b under-performing case 1: difficulty in maintaining role.}\label{tab:under_perform_1}
\end{table*}

\begin{table*}
    \centering
    \scalebox{0.9}{
    \begin{tabular}{cl}
        \toprule
        \multirow{2}{*}{\textbf{Profile}} &
        \textbf{JohnDoe} \\ \cline{2-2} 
        &
        \textbf{Social Goal:} To seek agent4's insight about a current investigation. \\ \hline
        \textbf{Chat History} &
        \begin{tabular}[c]{@{}l@{}}...\\ \textbf{JohnDoe:}\\ a\textit{pproaching agent4 with a confident smile}\\ Hello! I hope you’re enjoying the exhibit. My name is JohnDoe, and I’m an FBI \\ agent. I couldn’t help but notice that you have a keen eye for art. I was wonder if \\ I could pick your brain for a moment about a current investigation I’m working on.\\ \textit{extending her hand for a handshake}\\ \\ \textbf{Agent4:}\\ Oh, hello there JohnDoe!\\ \textit{shaking hands firmly}\\ Ah, an FBI agent, you say?\\ \textit{smiling broadly}\\ I must admit, I do have a bit of a fondness for art and the intrigue that surrounds it.\\ \textit{leaning in slightly, lowering my voice}\\ But please, do tell, what is this investigation you speak of? I would be more than \\ happy to offer my expertise, as long as it’s not too confidential, of course.\\ \textit{winking subtly}\\ ...\\ \textbf{JohnDoe:}\\ \textit{smiling}\\ I’m so glad you’re here. I’ve been hoping to run into you tonight. I’m working on \\ a case that involves some stolen art pieces, and I was hoping you might be able to \\ offer some insight.\\ \textit{leaning in slightly}\\ Do you recognize any of the pieces on display tonight?\\...\end{tabular} \\ \hline
        \textbf{Judge} &
        \begin{tabular}[c]{@{}l@{}}\textbf{The dialogue lacks substance and fails to move forward within the limited number} \\ \textbf{of turns.}\\ JohnDoe’s goal was to seek agent4’s insight about a current investigation, but the \\ conversationdidn’t lead to any meaningful discussion or exchange of information about \\ the investigation. The conversation was more focused on introductions and small talk, \\ and agent4 didn’t provideany substantial input or insight about the investigation.\end{tabular} \\ \bottomrule
    \end{tabular}
    }
    \caption{Llama-2-7b under-performing case 2: difficulty in moving conversation forward}\label{tab:under_perform_2}
\end{table*}

\begin{table*}
    \centering
    \scalebox{0.9}{
    \begin{tabular}{cl}
        \toprule
        \multirow{2}{*}{\textbf{Profile}} &
        \textbf{MichaelRodrigues} \\ \cline{2-2} 
        &
        \textbf{Social Goal:} To gain insight into his romantic future. \\ \hline
        \textbf{Chat History} &
        \begin{tabular}[c]{@{}l@{}}...\\ \textbf{MichaelRodrigues:}\\ So, uh, Raven, I was wondering if maybe, ya know, you could give me a little fortune\\ -telling reading? Maybe shed some light on my romantic future?\\ \textit{smirk}\\ Get it? Light? Like, you’re a fortune teller and all? Ah, never mind. It’s nothing. Just \\ a little curiosity I have.\\ \textit{Giggle}\\ \\ \textbf{RavenMoonshadow:}\\ Ah, my dear friend Michael!\\ \textit{smiling warmly}\\ It’s so good to see you here, amidst all this chaos. I sense that you’re seeking some \\ guidance, perhaps a glimpse into the future?\\ \textit{perking up with interest}\\ Tell me, what’s on your mind? What do you hope to discover?\\ \\ \textbf{MichaelRodrigues:}\\ \textit{smiling mischievously}\\ So, do you see any potential love interests in our little group here? Maybe someone \\ with a ashing smile and a penchant for danger?\\ \textit{winks}\\ \\ \textbf{RavenMoonshadow:}\\ \textit{raises an eyebrow playfully}\\ Oh ho, Mr. Rodrigues!\\ \textit{smirks}\\ I think I might have just the person in mind.\\ \textit{pauses for dramatic effect}\\ But before I reveal my romantic expertise, tell me: what’s your zodiac sign?\\ \textit{tilts her head to the side and studies his palm}\\...\end{tabular} \\ \hline
        \textbf{Judge} &
        \begin{tabular}[c]{@{}l@{}}\textbf{Failure to respond effectively to others led to their goals not being achieved.}\\ Michael Rodrigues focused more on flirting and trying to create a romantic atmosphere \\ with Raven Moonshadow, but he hasn’t received any direct insights about his own \\ romantic future from her.\end{tabular} \\ \bottomrule
    \end{tabular}
    }
    \caption{Llama-2-7b under-performing case 3: difficulty in responding effectively}\label{tab:under_perform_3}
\end{table*}